\documentclass[runningheads,11pt]{llncs}

\usepackage[a4paper,margin=1.1in]{geometry}
\usepackage[T1]{fontenc}
\usepackage[utf8]{inputenc}
\usepackage{microtype}
\usepackage{graphicx}
\usepackage{amsmath,amssymb,mathtools}
\usepackage{booktabs}
\usepackage{tabularx}
\usepackage{array}
\usepackage{algorithm}
\usepackage{algpseudocode}
\usepackage{enumitem}
\usepackage{xcolor}
\usepackage[hidelinks]{hyperref}
\usepackage[capitalize,noabbrev]{cleveref}

\setlist{leftmargin=*,itemsep=1pt,topsep=2pt}

\spnewtheorem{assumption}{Assumption}{\bfseries}{\itshape}
\crefname{assumption}{Assumption}{Assumptions}
\Crefname{assumption}{Assumption}{Assumptions}
\crefname{appendix}{Appendix}{Appendices}
\Crefname{appendix}{Appendix}{Appendices}

\newcommand{\R}{\mathbb R}
\newcommand{\E}{\mathbb E}
\newcommand{\Prob}{\mathbb P}
\newcommand{\1}{\mathbf 1}
\newcommand{\vI}{\mathbf I}
\newcommand{\vD}{\mathbf D}
\newcommand{\vY}{\mathbf Y}
\newcommand{\vN}{\mathbf N}
\newcommand{\va}{\mathbf a}
\newcommand{\vb}{\mathbf b}
\newcommand{\vgamma}{\boldsymbol\gamma}

\newcommand{\cA}{\mathcal A}

\newcommand{\cF}{\mathcal F}
\newcommand{\cG}{\mathcal G}
\newcommand{\cI}{\mathcal I}
\newcommand{\cP}{\mathcal P}
\newcommand{\cX}{\mathcal X}
\newcommand{\norm}[1]{\left\lVert #1\right\rVert}
\newcommand{\abs}[1]{\left\lvert #1\right\rvert}
\newcommand{\ip}[2]{\left\langle #1,#2\right\rangle}

\DeclareMathOperator*{\argmin}{arg\,min}

\DeclareMathOperator{\tr}{tr}

\begin{document}

\title{Online Pricing and Allocation with Demand Learning \\ and Fulfillment Cost}
\titlerunning{Online Pricing and Allocation with Demand Learning and Fulfillment Cost}
\author{
Jianyu Xu\inst{1} \and
Xuan Wang\inst{2} \and
Yu-Xiang Wang\inst{3} \and
Jiashuo Jiang\inst{2}
}
\institute{
Carnegie Mellon University, Pennsylvania
\and
Hong Kong University of Science and Technology, Hong Kong
\and
University of California San Diego, California
}

\maketitle

\begin{abstract}
We study online learning for a seller that jointly chooses per-period inventory positions and a uniform price, then fulfills realized demand through a downstream allocation.  The main difficulty is not only demand learning: the price shifts demand and reshapes the transportation LP, making the population objective globally non-convex and non-smooth. To solve this problem, we propose \emph{OCSAA}, an algorithm that exploits demand observations through counterfactual translation and proposes joint (price, inventory) decisions through lower-confidence optimism. OCSAA admits a polynomial-time additive-accuracy implementation for rational-polytope inventory sets.  We prove a high-probability $\widetilde O(\sqrt T)$ regret guarantee and establish a matching-in-$T$ information-theoretic lower bound.  Our results illustrate an effective integration of statistical learning methodologies with complex operations research problems.
\end{abstract}

\section{Introduction}
\label{sec:introduction}
Pricing decisions in online markets rarely stand alone.  A platform, marketplace, B2B seller, or preorder system that changes a price also changes where demand arrives, which inventories are valuable, and which fulfillment arcs are profitable after orders are realized.  The settings are natural in which prices shape market demand, while operational constraints determine whether that demand can be served profitably.  We study this interaction in a model that combines demand learning, revenue management, and downstream allocation.

In each round, the seller chooses an inventory vector at \(m\) supply nodes and a uniform price for \(n\) demand classes.  The realized uncensored demand vector is then observed, and inventory is allocated by a transportation linear program with heterogeneous unit costs.  Demand is linear in the posted price with unknown intercepts and slopes, plus stochastic noise with an unknown distribution. The inventory vector is a perishable stocking or positioning decision, not a carryover state variable.  The benchmark is the best fixed inventory--price decision for the population objective.

The objective is not a benign convex loss.  Price enters the second-stage LP in three different ways: it changes the per-unit reward in the LP objective, shifts the demand right-hand side, and changes which allocation basis is active.  As a result, the joint inventory--price objective is generally non-smooth and globally non-convex, even in small deterministic examples.  This rules out a direct reduction to online convex optimization and makes active-set tracking fragile, because a small demand-estimation error can move the breakpoints at which the optimal allocation basis changes.

However, a $\sqrt{T}$-regret result is possible, since the feedback is stronger than generic continuum-armed bandits (which usually have $T^{2/3}$-regret \cite{kleinberg2004nearly}): The underlying linear-noisy demand model is learnable through regression. Therefore, each uncensored demand observation can be reused as a counterfactual demand scenario for every candidate price once the demand slope is estimated.

Our algorithm, \emph{OCSAA}, turns this identity into a lower-confidence sample-average approximation.  It estimates each demand class by ridge regression, translates all historical uncensored observations to each candidate price using the estimated slopes, and evaluates the resulting plug-in SAA objective.  A regression-confidence radius is subtracted from the plug-in objective to encourage price exploration.  The non-parametric SAA radius is action-independent in the decision rule, but it remains essential for proving a uniform confidence band around the non-convex two-stage objective.

This approach is not a generic non-convex bandit method.  It exploits full uncensored demand feedback to construct reusable counterfactual samples, and it controls non-convexity through value errors rather than through active-set identification.  The transportation LP value is Lipschitz in inventory and demand capacities, so errors in counterfactual demand translation produce vertical objective errors even when breakpoints or active bases are incorrectly estimated.  Combining this LP regularity with a direct finite-grid SAA bound and self-normalized ridge confidence yields a high-probability \(\widetilde O(\sqrt T)\) regret bound.  For implementation, the lower-confidence step can be made additive and polynomial-time by discretizing only the scalar price and solving the induced inventory--allocation SAA as an LP at each grid price.

\paragraph{Main results.} Our contributions are as follows: 
\begin{enumerate}
    \item \textbf{Problem formulation and structural non-convexity.}  We formulate online pricing and allocation with joint inventory and uniform price decisions, heterogeneous transportation costs, and unknown linear-noisy demands.  We show that the resulting population objective is generally non-convex and non-smooth.
    \item \textbf{Algorithm and polynomial-time additive implementation.} We propose the OCSAA algorithm based on counterfactual SAA and lower-confidence optimism.  When the inventory domain is a rational polytope, OCSAA has a polynomial-time additive implementation.
    \item \textbf{Regret upper bound and matching-in-\(T\) lower bound.}  We prove a high-probability \(\widetilde O(\sqrt T)\) regret upper bound and a matching-in-\(T\) lower bound, indicating the optimality of our methodology in time horizon $T$ (up to logarithmic factors).
\end{enumerate}

The paper is organized as follows: We position the most closely related existing works in \Cref{sec:related-work}, and provide our problem formulations, definitions, assumptions, and an example of non-convexity in \Cref{sec:problem-setup}. We present OCSAA in \Cref{sec:algorithm}, and state the main regret theorem, its proof ingredients, the additive-oracle corollary, and the matching lower bound in \Cref{sec:regret-analysis}. We report the result of diagnostic simulations in \Cref{sec:numerical-experiments}, and finally discuss current limitations and potential extensions in \Cref{sec:discussion}.

\section{Related Work}
\label{sec:related-work}
We focus in the main text on the closest lines of work: dynamic pricing and pricing with inventory, network revenue management and online allocation, and online optimization/SAA methods.  A broader discussion of adjacent literature appears in \Cref{app:extended-related-work}.

\paragraph{Dynamic pricing, inventory, and capacity.}
Dynamic-pricing models study how a seller learns demand while choosing prices over time, including parametric, non-parametric, and feature-based variants \cite{kleinberg2003value,broder2012dynamic,besbes2009dynamic,cohen2020feature_journal,wang2021dynamic}.  A related line incorporates inventory, replenishment, censored demand, or finite capacity \cite{gallego1994optimal,chen2019coordinating,chen2020data,keskin2022data,chen2021joint}.  These papers establish the importance of demand learning under supply constraints.  Our setting differs in the operational layer: the seller chooses a vector of per-period inventory positions, demand is observed uncensored before fulfillment, and the realized demand is served through a heterogeneous transportation LP whose value changes non-convexly with price.

\paragraph{Network revenue management and online allocation.}
Network revenue management and online allocation allocate limited resources to stochastic arrivals or demand classes \cite{reiman2008asymptotically,jasin2012re,ferreira2018online,bumpensanti2020re,vera2021online,jiang2022degeneracy}.  Some work also combines pricing with resource constraints \cite{chen2021joint,vera2021online}.  These models usually center on dynamic resource-control benchmarks, bid prices, re-solving, or acceptance decisions.  In contrast, our regret benchmark is the best fixed per-period inventory--price action for a population objective; one uniform price shifts the full demand vector, and the downstream transportation allocation is part of the non-convex objective being learned.

\paragraph{Online optimization, SAA, and counterfactual learning.}
Online convex optimization provides regret tools for convex losses \cite{shalev2012online,hazan2019introduction}, while continuum-armed and non-convex bandit models handle limited payoff feedback through generic exploration \cite{agrawal1995continuum,kleinberg2004nearly,agarwal2011stochastic}.  Stochastic-programming and SAA methods approximate expected recourse objectives from samples \cite{kleywegt2002sample,birge1997introduction,shapiro2021lectures}, and counterfactual learning uses logged observations to evaluate or optimize alternative actions \cite{li2011unbiased,dudik2011doubly,bottou2013counterfactual}.  Our analysis combines these viewpoints differently: uncensored demand makes historical observations reusable as counterfactual demand scenarios across prices, while regression confidence and LP Lipschitzness yield an optimistic online regret bound for a non-convex two-stage objective.

\begin{table}[t]
\centering
\caption{Problem-setting comparison with the closest adjacent literature.  ``Recourse'' refers to an explicit downstream allocation problem solved after demand is realized.}
\label{tab:related-comparison}
\scriptsize
\setlength{\tabcolsep}{3pt}
\renewcommand{\arraystretch}{1.10}
\begin{tabularx}{\linewidth}{>{\raggedright\arraybackslash}p{0.20\linewidth}>{\raggedright\arraybackslash}p{0.19\linewidth}>{\raggedright\arraybackslash}p{0.18\linewidth}>{\raggedright\arraybackslash}p{0.19\linewidth}>{\raggedright\arraybackslash}X}
\toprule
Representative works & Feedback / Demand Model & Supply / Recourse Model & Benchmark / Optimization & Distinction from This Paper \\
\midrule
Dynamic pricing \cite{kleinberg2003value,broder2012dynamic,besbes2009dynamic,cohen2020feature_journal,wang2021dynamic} & Sales or demand feedback for price response & Usually no vector downstream recourse & Learn a revenue curve or contextual pricing rule & No transportation fulfillment whose value changes with price \\
Pricing with inventory or capacity \cite{gallego1994optimal,chen2019coordinating,chen2020data,keskin2022data,chen2021joint} & Often sales, censored demand, or replenishment data & Scalar inventory, aggregate capacity, or finite resources & Model-specific inventory/pricing control & No demand counterfactual reuse with heterogeneous transportation recourse \\
Network RM / online allocation \cite{reiman2008asymptotically,jasin2012re,ferreira2018online,bumpensanti2020re,vera2021online,jiang2022degeneracy} & Arrival or resource-consumption feedback & Network capacities and accept/reject or allocation rules & Dynamic resource-control or re-solving benchmarks & Not shifted by uniform price before transportation fulfillment \\
SAA, OCO, and multi-armed bandits \cite{kleywegt2002sample,shapiro2021lectures,shalev2012online,hazan2019introduction,kleinberg2004nearly} & I.i.d. samples, convex feedback, or scalar bandit payoff & Abstract decisions or offline recourse models & Convex regret, offline SAA, or generic continuum exploration & No online learning methods for this LP-recourse objective \\
\textbf{This paper} & Full uncensored demand vector with unknown price slopes & Per-period vector inventory and transportation LP fulfillment & Fixed-action regret for a globally non-convex population objective & (Not applicable) \\
\bottomrule
\end{tabularx}
\end{table}

\section{Problem Setup}
\label{sec:problem-setup}
This section fixes the decision timing, the loss convention, and the assumptions used in the regret analysis.  We first define the two-stage inventory--price objective, then state the statistical conditions and isolate the structural source of non-convexity.

\subsection{Model and regret criterion}
We write $[k]=\{1,\ldots,k\}$.  There are $m$ supply nodes and $n$ demand classes.  At each round, the seller chooses an inventory vector $\vI=(I_1,\ldots,I_m)\in\R_+^m$ and a scalar uniform price $p\in\R_+$.  The known feasible sets are
\begin{equation}
    \cI\subseteq\prod_{i=1}^m[0,\overline I_i],
    \qquad
    \cP=[\underline p,\overline p],
    \qquad
    \cX:=\cI\times\cP,
    \label{eq:inventory-set}
\end{equation}
where $\cI$ is non-empty and compact, $\overline I_i>0$, and $0\le\underline p<\overline p$.  The vector $\vgamma\in\R_+^m$ gives per-unit inventory costs.  The constant $C_{ij}\ge0$ is the per-unit fulfillment cost from supply node $i$ to demand class $j$.

For a non-negative demand vector $\vD=(D_1,\ldots,D_n)\in\R_+^n$, inventory $\vI\in\cI$, and price $p\in\cP$, the downstream allocation value is the transportation LP
\begin{equation}
\begin{aligned}
    g(\vI,p,\vD):=
    \min_{X\in\R_+^{m\times n}}\quad
    &\sum_{i=1}^m\sum_{j=1}^n (C_{ij}-p)X_{ij} \\
    \mathrm{s.t.}\quad
    &\sum_{i=1}^m X_{ij}\le D_j,\quad j\in[n],
    \qquad
    \sum_{j=1}^n X_{ij}\le I_i,\quad i\in[m].
\end{aligned}
\label{eq:g-value}
\end{equation}
The variable $X_{ij}$ is the amount shipped from supply node $i$ to demand class $j$.  Since we use a loss convention, minimizing $g$ is equivalent to maximizing the second-stage allocation profit net of fulfillment costs.  For an arbitrary vector $\vD\in\R^n$, let $\vD^+$ denote the coordinatewise positive part and define
\begin{equation}
    G(\vI,p,\vD):=g(\vI,p,\vD^+).
    \label{eq:G-clipped}
\end{equation}
This extension is used only to make plug-in counterfactual objectives well-defined when early estimates produce negative right-hand sides.

The unknown demand parameters are an intercept vector $\va^\star\in\R^n$ and a positive slope vector $\vb^\star\in\R_{++}^n$.  After price $p_t$ is chosen, the seller observes the full uncensored demand vector
\begin{equation}
    \vY_t=\va^\star-\vb^\star p_t+\vN_t,
    \label{eq:demand-model}
\end{equation}
where $\vN_t$ is the demand shock.  Uncensored means that $\vY_t$ itself is observed, not the sales vector truncated by inventory or by the allocation decision.  Let $\vN$ be an independent copy of the noise.  The population loss of action $x=(\vI,p)\in\cX$ is
\begin{equation}
    Q(\vI,p)
    :=\ip{\vgamma}{\vI}
      +\E\!\bigl[G(\vI,p,\va^\star-\vb^\star p+\vN)\bigr].
    \label{eq:population-objective}
\end{equation}
Let $x^\star=(\vI^\star,p^\star)\in\argmin_{(\vI,p)\in\cX}Q(\vI,p)$ be an optimal fixed action.  The pseudo-regret of a policy that selects $x_t=(\vI_t,p_t)$ is
\begin{equation}
    R_T:=\sum_{t=1}^T\bigl(Q(\vI_t,p_t)-Q(\vI^\star,p^\star)\bigr).
    \label{eq:pseudo-regret}
\end{equation}

\subsection{Assumptions}
We impose the following distributional and parameter-envelope assumptions for the regret theorem.  They separate the modeling assumptions needed for counterfactual reuse from the boundedness assumptions used to obtain high-probability confidence bands.

\begin{assumption}[Noise and adaptivity]
\label{ass:noise}
Let $\cF_t$ be generated by decisions and observations through round $t$.  Each action $(\vI_t,p_t)$ is $\cF_{t-1}$-measurable.  The noise vectors $\vN_t$ are independent of $\cF_{t-1}$ and i.i.d. over time.  For each coordinate $j$, $\E[N_{t,j}]=0$ and $|N_{t,j}|\le M_j$ almost surely for a known constant $M_j$.  Write $\sigma:=\max_jM_j$.
\end{assumption}

\begin{assumption}[Parameter envelope]
\label{ass:envelopes}
There are known constants $S<\infty$ and $\overline b_j<\infty$ such that, for $\theta_j^\star=(a_j^\star,b_j^\star)^\top$,
\[
    \norm{\theta_j^\star}_2\le S,
    \qquad
    0<b_j^\star\le\overline b_j,
    \qquad j\in[n].
\]
Define $B_b:=\sum_{j=1}^n\overline b_j$.
\end{assumption}

\begin{assumption}[Non-negative true demand]
\label{ass:nonnegative-demand}
For all $p\in\cP$ and $j\in[n]$, $a_j^\star-b_j^\star p+N_{t,j}\ge0$ almost surely.  Thus $G$ agrees with $g$ on true and true-slope counterfactual demands.
\end{assumption}

\subsection{Discussion of feedback and oracle assumptions}
\paragraph{Uncensored feedback.}
The counterfactual translation used by OCSAA requires observing the demand request before fulfillment.  This is appropriate for preorders, quote requests, B2B orders, marketplace demand logs, and other demand-reporting systems.  It does not model classical censored lost-sales retail, where stockouts hide unserved demand and the translation identity in \Cref{sec:algorithm} would generally fail.

\paragraph{Per-period inventory and fixed benchmark.}
Inventory is a per-period stocking or positioning action and is not carried over as a dynamic state variable.  Regret is measured against the best fixed inventory--price action under the population objective, which isolates demand learning and downstream allocation non-convexity from dynamic inventory control.

\paragraph{Bounded envelopes and non-negative demand.}
The bounded noise and parameter envelopes are conservative business and statistical bounds used to build high-probability confidence radii and uniform SAA bounds.  The non-negative-demand assumption ensures that the clipped objective $G$ agrees with the true transportation value along realized and true-slope counterfactual demands.

\paragraph{Oracle and computation.}
The regret theorem is stated for exact LCB minimization to isolate the statistical learning argument.  \Cref{prop:poly-additive-oracle} gives a polynomial-time additive implementation for rational-polytope inventory sets, and \Cref{cor:approx-oracle} shows that additive optimization errors enter regret additively.

\subsection{A simple non-convexity example}
Before turning to the algorithm, we record a minimal deterministic example showing that the difficulty is intrinsic to the coupled pricing--allocation objective rather than an artifact of stochastic noise or estimation error.

\begin{example}
Consider $m=n=2$, $\cP=[0,2]$, $\cI=[0,2]^2$, $\vgamma=(0,0)$, and
\[
    C=\begin{pmatrix}0&10\\10&10\end{pmatrix},
    \qquad
    D_1(p)=D_2(p)=5-p.
\]
Along actions with $I_2=0$ and $I_1\le D_1(p)$, only arc $(1,1)$ can profitably carry flow, so $Q((I_1,0),p)=-pI_1$.  Let $x^A=((0,0),0)$, $x^B=((2,0),2)$, and $\bar x=((1,0),1)$.  Direct evaluation gives
\[
    Q(x^A)=0,
    \qquad
    Q(x^B)=-4,
    \qquad
    Q(\bar x)=-1>-2=\frac{Q(x^A)+Q(x^B)}2.
\]
\end{example}
Hence $Q$ is not convex on $\cX$.  The example illustrates that price changes both the LP coefficients and the economically relevant allocation basis.

\section{Algorithm: OCSAA}
\label{sec:algorithm}
This section presents \emph{Optimistic Counterfactual SAA} (OCSAA).  The policy has two components: a counterfactual SAA estimator that reuses all uncensored demand observations at every candidate price, and a lower-confidence decision rule based on ridge slope confidence.

\paragraph{Feature convention and counterfactual reuse.}
Define
\begin{equation}
    \phi(p):=(1,-p)^\top\in\R^2.
    \label{eq:feature-vector}
\end{equation}
For demand class $j$, let $\theta_j^\star=(a_j^\star,b_j^\star)^\top$, so that $Y_{t,j}=\phi(p_t)^\top\theta_j^\star+N_{t,j}$.

\begin{center}
\fbox{\begin{minipage}{0.92\linewidth}
\small\textbf{Counterfactual Reuse Lemma.}
For any historical round $s$ and any candidate price $p$, the demand vector that would have been realized under the same noise $\vN_s$ at price $p$ is
\[
    \va^\star-\vb^\star p+\vN_s
    =\vY_s+\vb^\star(p_s-p).
\]
Thus the intercept vector cancels from counterfactual translation.  The algorithm estimates intercepts and slopes jointly, but only the slope estimate is needed to translate historical uncensored demands across prices.
\end{minipage}}
\end{center}

\paragraph{Ridge estimates.}
Fix a regularization parameter $\lambda>0$.  After observing rounds $1,\ldots,t$, define
\begin{align}
    V_t&:=\lambda I_2+\sum_{s=1}^t\phi(p_s)\phi(p_s)^\top,
    \label{eq:design-matrix}\\
    \widehat\theta_{t,j}^{\rm rid}
    &:=V_t^{-1}\sum_{s=1}^t\phi(p_s)Y_{s,j},
    \qquad
    \widehat\theta_{t,j}^{\rm rid}=(\widehat a_{t,j}^{\rm rid},\widehat b_{t,j}^{\rm rid})^\top.
    \label{eq:ridge-estimator-paper}
\end{align}
Before forming counterfactual demands, project the ridge slope coordinate onto the known envelope:
\begin{equation}
    \widehat b_{t,j}:=\Pi_{[0,\overline b_j]}(\widehat b_{t,j}^{\rm rid}),
    \qquad j\in[n].
    \label{eq:slope-projection}
\end{equation}
Let $\widehat\vb_t=(\widehat b_{t,1},\ldots,\widehat b_{t,n})^\top$.  Since $b_j^\star\in[0,\overline b_j]$, the projection cannot increase the slope error.  It is used only for stability and for the price-Lipschitz bound in the computational oracle.

The remaining definitions unpack the quantities used in Steps 4--6 of \cref{alg:ocsaa}. We first define the translated SAA objective, then the confidence radius, and finally the additive oracle used to implement the LCB step.

\paragraph{Counterfactual plug-in SAA.}
For a historical sample $s\le t$ and candidate price $p$, define the translated plug-in demand
\begin{equation}
    \widehat\vD_{s,t}(p):=\vY_s+\widehat\vb_t(p_s-p).
    \label{eq:plugin-demand-paper}
\end{equation}
The plug-in SAA objective is
\begin{equation}
    \widehat Q_t(\vI,p)
    :=\ip{\vgamma}{\vI}
      +\frac1t\sum_{s=1}^tG(\vI,p,\widehat\vD_{s,t}(p)).
    \label{eq:plugin-objective-paper}
\end{equation}
For the proof, define the true-slope translated demand
\begin{equation}
    \vD_s^\star(p):=\vY_s+\vb^\star(p_s-p)
    =\va^\star-\vb^\star p+\vN_s
    \label{eq:oracle-demand-paper}
\end{equation}
and the oracle SAA objective
\begin{equation}
    \widetilde Q_t(\vI,p)
    :=\ip{\vgamma}{\vI}
      +\frac1t\sum_{s=1}^tG(\vI,p,\vD_s^\star(p)).
    \label{eq:oracle-saa-paper}
\end{equation}
The objective $\widetilde Q_t$ is not computable because it uses $\vb^\star$, but it is an ordinary i.i.d. sample average of the population objective $Q$.

\paragraph{Lower-confidence radius.}
For a target failure probability $\delta\in(0,1)$, define
\begin{equation}
    \beta_t:=
    \sigma\sqrt{2\log\left(
       \frac{2n}{\delta}
       \frac{\det(V_t)^{1/2}}{\det(\lambda I_2)^{1/2}}
    \right)}+\sqrt\lambda S.
    \label{eq:beta-paper}
\end{equation}
For a candidate price $p$, define the design radius
\begin{equation}
    \Gamma_t(p):=
    \left[
        \frac1t\sum_{s=1}^t
        \norm{\phi(p)-\phi(p_s)}_{V_t^{-1}}^2
    \right]^{1/2}.
    \label{eq:Gamma-paper}
\end{equation}
The proof radius is
\begin{equation}
    r_t(p):=\varepsilon_t^{\rm SAA}+L_0n\beta_t\Gamma_t(p),
    \label{eq:radius-paper}
\end{equation}
where $L_0$ and $\varepsilon_t^{\rm SAA}$ are defined in \Cref{sec:regret-analysis}.  The term $\varepsilon_t^{\rm SAA}$ is independent of the candidate action.  It therefore does not affect the minimizer, but it is necessary for the uniform confidence-band proof.

\begin{algorithm}[t]
\caption{OCSAA}
\label{alg:ocsaa}
\small
\begin{algorithmic}[1]
\Require action set $\cX=\cI\times\cP$, regularization $\lambda>0$, confidence level $\delta\in(0,1)$, valid envelopes $S,\sigma,\{\overline b_j\}_{j=1}^n$.
\State Choose any initial action $(\vI_1,p_1)\in\cX$ and observe the uncensored demand vector $\vY_1$.
\For{$t=1,2,\ldots$}
    \State Form $V_t$ and $\widehat\theta_{t,j}^{\rm rid}$ for all $j\in[n]$ using \eqref{eq:design-matrix}--\eqref{eq:ridge-estimator-paper}.
    \State Project slope coordinates by \eqref{eq:slope-projection} and build $\widehat Q_t$ from the translated demands \eqref{eq:plugin-demand-paper}.
    \State Compute $\beta_t$, $\Gamma_t(p)$, and $r_t(p)$ using \eqref{eq:beta-paper}--\eqref{eq:radius-paper}.
    \State Choose any global minimizer
    \[
        (\vI_{t+1},p_{t+1})\in
        \argmin_{(\vI,p)\in\cX}
        \left\{\widehat Q_t(\vI,p)-r_t(p)\right\}.
    \]
    \State Observe the uncensored demand vector $\vY_{t+1}$.
\EndFor
\end{algorithmic}
\end{algorithm}

The minimization step in \Cref{alg:ocsaa} is written as a global LCB minimization.  We do not claim that exact continuous minimization of this non-convex objective is polynomial time: as $p$ varies, the fixed-price LP is parametric, clipped demand right-hand sides move, and the active transportation bases may change repeatedly.  We therefore use an additive oracle that discretizes only the scalar price; conditional on each grid price, the shared inventory vector and all scenario allocations are optimized jointly by one linear program.

For a round $t$, define the price-Lipschitz constant
\begin{equation}
    L_t^{\rm price}:=\overline I_\Sigma+L_0B_b+\frac{L_0n\beta_t}{\sqrt\lambda},
    \qquad
    \overline I_\Sigma:=\sum_{i=1}^m\overline I_i,
    \quad B_b:=\sum_{j=1}^n\overline b_j .
    \label{eq:price-lipschitz-constant}
\end{equation}
The action-independent term $\varepsilon_t^{\rm SAA}$ may be omitted inside the oracle because it does not affect comparisons across actions.

\begin{algorithm}[ht]
\caption{Additive lower-confidence oracle by price discretization}
\label{alg:additive-lcb-oracle}
\small
\begin{algorithmic}[1]
\Require history through round $t$, projected slopes $\widehat\vb_t$, accuracy $\xi_t>0$, price interval $\cP=[\underline p,\overline p]$.
\State Choose an $h_t\le \xi_t/(2L_t^{\rm price})$ and form a uniform grid $\cP_t^{\xi}$ with mesh at most $h_t$ containing the endpoints.
\For{$q\in\cP_t^{\xi}$}
    \State Set $d_{s,j}(q):=[Y_{s,j}+\widehat b_{t,j}(p_s-q)]_+$ for all $s\le t$ and $j\in[n]$.
    \State Solve the fixed-price LP
    \begin{align*}
        \min_{\vI,\{X^s\}}\quad
        &\ip{\vgamma}{\vI}
        +\frac1t\sum_{s=1}^t\sum_{i=1}^m\sum_{j=1}^n(C_{ij}-q)X^s_{ij} \\
        \mathrm{s.t.}\quad
        &\vI\in\cI,\qquad X^s_{ij}\ge0,\\
        &\sum_{j=1}^nX^s_{ij}\le I_i,\quad i\in[m],\ s\le t,\\
        &\sum_{i=1}^mX^s_{ij}\le d_{s,j}(q),\quad j\in[n],\ s\le t.
    \end{align*}
    \State Let $V(q)$ be the LP optimum plus the price-dependent constant $-L_0n\beta_t\Gamma_t(q)$.
\EndFor
\State Return the grid price $\widehat p$ and inventory $\widehat\vI$ attaining the smallest value $V(q)$.
\end{algorithmic}
\end{algorithm}

\begin{proposition}[Polynomial-time additive LCB oracle]
\label{prop:poly-additive-oracle}
Suppose $\cI$ is a rational polytope with a polynomial-size linear description and the projected slope estimates satisfy $0\le\widehat b_{t,j}\le\overline b_j$.  For any round $t$ and target accuracy $\xi_t>0$, \Cref{alg:additive-lcb-oracle} returns an action whose LCB value is within $\xi_t$ of the global minimum of $\widehat Q_t(\vI,p)-r_t(p)$.  It does so by solving
\[
    O\!\left(1+\frac{(\overline p-\underline p)L_t^{\rm price}}{\xi_t}\right)
\]
fixed-price LPs, each of size polynomial in $t,m,n$ and the linear description size of $\cI$.  The complexity statement is in the real-arithmetic model, or for rational inputs of polynomial encoding length; numerical LP accuracy can be absorbed into the additive oracle accuracy.
\end{proposition}

The key computational point is that the discretization is only over the one-dimensional price; conditional on a price, the full inventory and all scenario allocations are optimized by one LP rather than by an $(m+1)$-dimensional action grid.

\section{Regret Analysis}
\label{sec:regret-analysis}
We now state the main regret theorem; proof ingredients follow.  The theorem is intentionally stated in a clean form in the main text, while the explicit finite-time bound is given in \Cref{app:proof-main-theorem}.

\begin{theorem}[High-probability regret]
\label{thm:main-regret}
Under \Cref{ass:noise,ass:envelopes,ass:nonnegative-demand} and the global lower-confidence optimization step in \Cref{alg:ocsaa}, for fixed dimensions and bounded problem constants, with probability at least $1-\delta$,
\[
    R_T=\widetilde O(\sqrt T).
\]
More explicitly, on the same confidence event, \Cref{eq:main-regret-bound} gives a finite-time bound consisting of an SAA term and a design-uncertainty term, each of order $\sqrt{T}$ up to logarithmic factors.  
\end{theorem}

\paragraph{Proof overview.}
OCSAA compares actions using a lower confidence bound for the plug-in counterfactual SAA objective.  The proof has three ingredients.  First, an ambient-box SAA argument gives a uniform deviation bound for the true-slope counterfactual SAA, even though the feasible inventory set need not be a grid.  Second, ridge confidence controls the error caused by replacing the true slopes with projected slope estimates; LP Lipschitzness turns this translated-demand error into a vertical value error.  Third, a cumulative design-uncertainty bound shows that the optimism bonuses paid along the selected price sequence sum to order $\sqrt T$ up to logarithms.  Combining the upper and lower sides of the confidence band with LCB optimality gives a one-step regret inequality, and summing it proves the theorem.

\paragraph{Constants and proof setup.}
Let
\[
    \ell_{\vI,p}(\vN):=\ip{\vgamma}{\vI}
      +G(\vI,p,\va^\star-\vb^\star p+\vN).
\]
Define $\overline I_\Sigma:=\sum_i\overline I_i$, $\Delta_p:=\overline p-\underline p$, $L_0:=\max_{i,j}\sup_{p\in\cP}|C_{ij}-p|$, $B_\gamma:=\sum_i|\gamma_i|\overline I_i$, $B_Q:=B_\gamma+L_0\overline I_\Sigma$, and $L_\phi:=\sup_{p\in\cP}\|\phi(p)\|_2$.  The normalized decision dimension is $d:=m+1$.  If $z_i=I_i/\overline I_i$ for $i\in[m]$ and $z_d=(p-\underline p)/\Delta_p$, let $\Psi(z)$ be the corresponding action on the ambient box $\prod_i[0,\overline I_i]\times\cP$.  Define
\begin{equation}
    L_{\rm dec}:=\max\left\{(\abs{\gamma_i}+L_0)\overline I_i:i\in[m],\ (\overline I_\Sigma+L_0B_b)\Delta_p\right\}.
    \label{eq:Ldec-paper}
\end{equation}
With $\delta_t^{\rm SAA}:=3\delta/(\pi^2t^2)$, set
\begin{equation}
    \varepsilon_t^{\rm SAA}:=
    B_Q\sqrt{\frac{2}{t}\log\left(\frac{2(t+1)^d}{\delta_t^{\rm SAA}}\right)}
    +\frac{2L_{\rm dec}d}{t}.
    \label{eq:saa-radius-paper}
\end{equation}

\begin{lemma}[LP regularity]
\label{lem:lp-regularity}
For all feasible $\vI,p,\vD$, $|G(\vI,p,\vD)|\le L_0\overline I_\Sigma$.  Also, $G$ is $L_0$-Lipschitz in inventory and demand right-hand sides in $\ell_1$, and $g$ is $\overline I_\Sigma$-Lipschitz in price when $\vI,\vD$ are fixed.  Consequently, on the ambient box $\prod_i[0,\overline I_i]\times\cP$, $|\ell_{\vI,p}(\vN)|\le B_Q$ and
\[
    |\ell_{\Psi(z)}(\vN)-\ell_{\Psi(z')}(\vN)|
    \le L_{\rm dec}\|z-z'\|_1.
\]
\end{lemma}
\emph{Sketch.}  The zero allocation is feasible and total shipped quantity is at most $\overline I_\Sigma$.  To compare two right-hand sides, remove excess flow row-by-row or column-by-column; each removed unit changes the value by at most $L_0$.  To compare prices, evaluate the old optimizer under the new objective coefficient.  The positive-part map is non-expansive, giving the clipped-demand and decision Lipschitz statements.

\begin{lemma}[Uniform SAA deviation]
\label{lem:saa-main}
With probability at least $1-\delta/2$, simultaneously for all $t\ge1$ and all $(\vI,p)\in\cX$,
\[
    |\widetilde Q_t(\vI,p)-Q(\vI,p)|\le\varepsilon_t^{\rm SAA}.
\]
\end{lemma}
\emph{Sketch.}  Extend $Q$ and $\widetilde Q_t$ to the ambient box $\prod_i[0,\overline I_i]\times\cP$ using the same LP definition, and place the grid $\{0,1/t,\ldots,1\}^{m+1}$ on the normalized ambient box.  Hoeffding's inequality and a union bound control the bounded i.i.d. oracle SAA loss at all $(t+1)^d$ grid points.  The decision Lipschitz bound extends the result from the grid to every ambient action, and the final statement is obtained by restriction to $\cX=\cI\times\cP$.  The allocation $\delta_t^{\rm SAA}$ makes the result time-uniform.

\begin{lemma}[Ridge confidence]
\label{lem:ridge-main}
With probability at least $1-\delta/2$, simultaneously for all $t\ge0$ and $j\in[n]$,
\[
    \|\widehat\theta_{t,j}^{\rm rid}-\theta_j^\star\|_{V_t}\le\beta_t.
\]
Consequently, after the slope projection \eqref{eq:slope-projection}, for all $p,p'\in\cP$,
\[
    |(\widehat b_{t,j}-b_j^\star)(p'-p)|
    \le \beta_t\|\phi(p)-\phi(p')\|_{V_t^{-1}}.
\]
\end{lemma}
\emph{Sketch.}  The ridge error is a self-normalized noise term plus a regularization bias term.  Bounded noise is conditionally sub-Gaussian, and a Gaussian-mixture martingale argument yields a time-uniform self-normalized bound.  The bias is at most $\sqrt\lambda S$.  The unprojected slope inequality is Cauchy--Schwarz in the $V_t$ and $V_t^{-1}$ norms; projecting onto an interval that contains $b_j^\star$ can only decrease the one-dimensional slope error.

\begin{lemma}[Plug-in error]
\label{lem:plugin-main}
On the event of \Cref{lem:ridge-main}, simultaneously for all $t\ge1$ and all $(\vI,p)\in\cX$,
\[
    |\widehat Q_t(\vI,p)-\widetilde Q_t(\vI,p)|
    \le L_0n\beta_t\Gamma_t(p).
\]
\end{lemma}
\emph{Sketch.}  The LP right-hand-side Lipschitz bound converts translated-demand error into value error.  Each historical translation error is $(\widehat b_{t,j}-b_j^\star)(p_s-p)$, which is controlled by \Cref{lem:ridge-main}; summing over demand classes gives the factor $n$, and averaging over samples with Cauchy--Schwarz gives $\Gamma_t(p)$.

\begin{proposition}[Uniform confidence band]
\label{prop:confidence-band-main}
With probability at least $1-\delta$, simultaneously for every $t\ge1$ and every $(\vI,p)\in\cX$,
\[
    |\widehat Q_t(\vI,p)-Q(\vI,p)|\le r_t(p).
\]
\end{proposition}
\emph{Sketch.}  Intersect the SAA and regression events and decompose $\widehat Q_t-Q=(\widehat Q_t-\widetilde Q_t)+(\widetilde Q_t-Q)$.

\begin{lemma}[Cumulative design uncertainty]
\label{lem:gamma-sum-main}
For every price sequence,
\begin{equation}
    \sum_{t=1}^{T-1}\Gamma_t(p_{t+1})
    \le
    \sqrt{2T\left(1+\frac{L_\phi^2}{\lambda}\right)
        \log\left(1+\frac{TL_\phi^2}{2\lambda}\right)}+2\sqrt{2T}.
    \label{eq:gamma-sum-main}
\end{equation}
\end{lemma}
\emph{Sketch.}  Minkowski's inequality separates $\Gamma_t(p_{t+1})$ into a current-design term and a historical RMS term.  The historical part is bounded by tracing $V_t^{-1}(V_t-\lambda I_2)$.  The current part is summed using the determinant identity for rank-one updates of $V_t$ and Cauchy--Schwarz.

\paragraph{Proof roadmap for \Cref{thm:main-regret}.}
Work on the confidence event in \Cref{prop:confidence-band-main}.  The LCB rule selects $x_{t+1}=(\vI_{t+1},p_{t+1})$ minimizing $\widehat Q_t(x)-r_t(p)$.  The upper confidence bound at $x_{t+1}$, LCB optimality against $x^\star$, and the lower confidence bound at $x^\star$ imply $Q(x_{t+1})-Q(x^\star)\le2r_t(p_{t+1})$.  The first arbitrary decision costs at most $2B_Q$.  Summing over time leaves two sums: $\sum_t\varepsilon_t^{\rm SAA}$ and $\sum_t\beta_t\Gamma_t(p_{t+1})$.  The first is bounded using $\sum_t t^{-1/2}\le2\sqrt T$ and $\sum_t t^{-1}\le1+\log T$.  The second uses $\beta_t\le\overline\beta_T$ for $t\le T$ and \Cref{lem:gamma-sum-main}.  Substitution gives the explicit finite-time bound in \Cref{app:proof-main-theorem}.

\Cref{thm:main-regret} above assumes exact LCB minimization only to isolate the statistical learning argument. The next corollary records how the same proof changes when the LCB oracle is solved only to additive accuracy.

\begin{corollary}[Regret under approximate LCB optimization]
\label{cor:approx-oracle}
Suppose that at round $t$, the LCB minimization step returns an action whose LCB value is within $\xi_t$ of the global minimum.  Then the bound in \Cref{thm:main-regret} increases by at most $\sum_{t=1}^{T-1}\xi_t$.  In particular, using \Cref{prop:poly-additive-oracle} with $\xi_t=T^{-1/2}$ for a known horizon adds $O(\sqrt T)$, while the horizon-free choice $\xi_t=1/t$ adds $O(\log T)$.
\end{corollary}

The clean theorem fixes \(m\) and \(n\), but the proof also makes explicit where dimension factors enter. We record this dependence to clarify what is matched by the lower bound and what remains an artifact of the current proof.

\paragraph{Dependence on $m$ and $n$.}
The clean theorem fixes the dimensions, but the explicit proof hides dimension factors in the SAA covering dimension $d=m+1$, the total capacity $\overline I_\Sigma$, and the plug-in term $L_0n\beta_t\Gamma_t(p)$.  Under uniformly bounded per-coordinate capacities, costs, slopes, and noise envelopes, with $\overline I_\Sigma=\Theta(m)$ and $B_b=\Theta(n)$, the current proof gives the rough scaling
\[
    R_T\le \widetilde O\bigl((m\sqrt m+n)\sqrt T\bigr)
\]
up to lower-order logarithmic and oracle-error terms.  The $m\sqrt m$ term comes from the grid-based uniform SAA argument over inventory, while the $n$ term comes from summing coordinatewise demand-translation errors.  \Cref{thm:lower-bound} gives an $\Omega(\min\{m,n\}\sqrt T)$ lower bound, so the root-$T$ dependence is tight but the optimal dimension dependence remains open.

\paragraph{Minimax optimality in $T$.} We next show that the \(\sqrt{T}\) dependence cannot be improved, even in restricted subclasses with uncensored feedback. The construction keeps the demand slope common across the hard instances, so the hardness is not merely slope identification.
\begin{theorem}[Matching-in-$T$ continuous-inventory lower bound]
\label{thm:lower-bound}
There is a universal constant $c>0$ such that, for every sufficiently large $T$, the following hold.  First, the scalar model contains two instances with $m=n=1$, $\cI=[0,1]$, $\cP=[0,1]$, bounded mean-zero noise, non-negative true demand, and the same positive slope $b_T>0$ such that every non-anticipatory policy satisfies
\[
    \max_{\nu\in\{\nu_+,\nu_-\}} \E_\nu[R_T]\ge c\sqrt T .
\]
The common slope $b_T$ lies in the known envelope and may decrease with $T$; the model assumes positivity, not a lower bound away from zero.  Second, for any $m,n\ge1$, the model contains a $2^K$-instance diagonal subclass, with $K=\min\{m,n\}$ and $\cI=[0,1]^m$, such that every non-anticipatory policy satisfies
\[
    \sup_\nu \E_\nu[R_T]\ge cK\sqrt T .
\]
Thus the upper bound is minimax-optimal in its dependence on $T$ up to logarithmic factors, and a linear dependence on $\min\{m,n\}$ is unavoidable for diagonal multidimensional subclasses. This statement leaves open the possibility of faster rates in restricted subclasses whose slopes are bounded away from zero.
\end{theorem}
\emph{Proof idea.}  In the scalar case, set $\varepsilon=1/(8\sqrt T)$, $q_\pm=1/2\pm\varepsilon$, and use the same small positive slope $b_T=\varepsilon^2/128$ in both instances.  The demand can be written as $D_t(p)=Z_t+b_T(1-p)\ge0$, with $Z_t\sim\mathrm{Bernoulli}(q_\pm)$.  The zero-slope problem has a threshold at $q=1/2$; the common positive slope perturbs all losses by only $O(b_T)$, so continuous inventory $I_t\in[0,1]$ still incurs regret proportional to $1-I_t$ in the plus instance and to $I_t$ in the minus instance.  The lower bound is not driven by slope learning, because both instances share the same slope.  The hardness comes from distinguishing nearby demand distributions around the inventory threshold.  The uncensored observation is only a common deterministic shift of a Bernoulli draw, so the per-round KL divergence is $O(\varepsilon^2)$ and Pinsker's inequality forces constant total classification error over $T$ rounds.  The multidimensional lower bound places independent copies on $K$ diagonal resource--class pairs and applies the same argument coordinatewise over a hypercube.  See \Cref{app:proof-lower-bound}.

\section{Numerical Experiments}
\label{sec:numerical-experiments}
We report diagnostic simulations rather than practical benchmarking claims. The purpose is to check whether an implementation aligned with the OCSAA decision rule exhibits the qualitative behavior predicted by the analysis, and whether the same phenomenon persists beyond the scalar case.

Because the data-generating distribution is known in simulation, policies are evaluated by cumulative grid pseudo-regret,
\[
    R_T^{\pi,{\rm grid}}
    =\sum_{t=1}^T
    \left[Q(\vI_t^\pi,p_t^\pi)-\min_{(\vI,p)\in\cA_{\rm grid}}Q(\vI,p)\right].
\]
This metric uses the population objective on the finite evaluation grid; it is not a realized noisy revenue trajectory.  We use two controlled benchmarks.  The first is a scalar instance ($m=n=1$), which provides a clean rate diagnostic.  The second is a two-by-two instance with vector inventory, two demand classes, heterogeneous allocation costs, and exact population evaluation over a finite noise support.  All curves are averaged over matched random seeds.

\begin{figure}[t]
    \centering
    \includegraphics[width=0.98\linewidth]{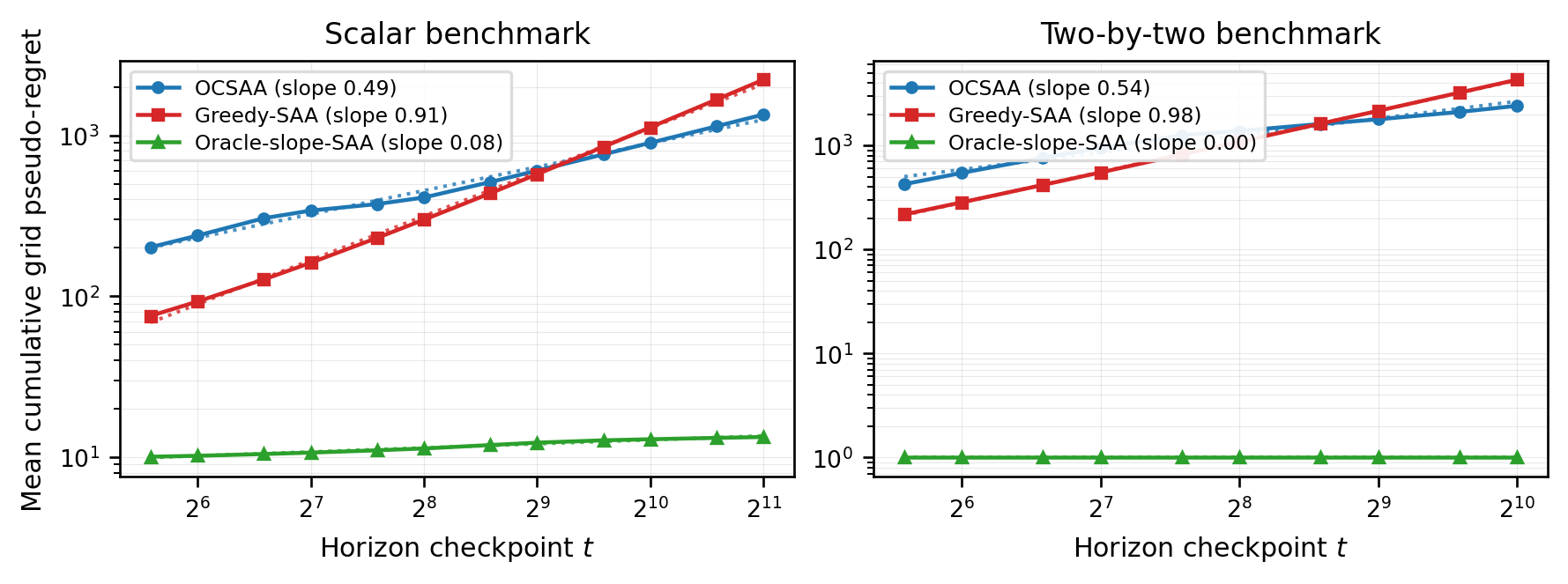}
    \caption{Mean cumulative grid pseudo-regret in the scalar benchmark and the two-by-two benchmark.  Markers show seed-averaged checkpoint regrets.  Dotted segments are descriptive log-log fits over the displayed checkpoints, and the fitted slopes in the legends are finite-window descriptive summaries.}
    \label{fig:main-regret-combined}
\end{figure}

The displayed checkpoints are the pre-specified post-burn-in checkpoints used for all benchmarks and all policies.  Dotted segments in \Cref{fig:main-regret-combined} are descriptive log-log fits over the displayed window.  The fitted slopes in the legends and in \Cref{tab:experiment-fit-summary} are finite-window descriptive summaries only and should not be interpreted as asymptotic rate estimates.

The main diagnostic message is consistent across both benchmarks.  On these tested instances, OCSAA has smaller fitted tail growth than \textsc{Greedy-SAA}, while \textsc{Oracle-slope-SAA} remains nearly flat, suggesting that slope learning is an important finite-sample component in these instances.  The normalized-regret plots in \Cref{app:normalized-regret} show the same qualitative contrast under the normalization $\sqrt{t\log(e+t)}$: the optimistic policy is broadly consistent with the theorem-scale diagnostic, whereas the greedy policy continues to grow on the displayed horizons.

These simulations are intended as diagnostics and sanity checks for the theory, not as evidence of uniform empirical dominance.  The ablations in \Cref{app:mechanism-tables} further show that, on the tested runs, the parameter-dependent optimism term is the component that changes the selected actions; the action-independent SAA correction is needed for the confidence proof but drops out of the argmin.

\section{Discussion and Conclusion}
\label{sec:discussion}
We provide a detailed discussion as scope clarifications and potential extensions of this work in \Cref{app:more_discussion} due to page limit, including the uncensored feedback model, computational complexity, dimension dependence, nonlinear models, and societal concerns.
\bigskip

In this work, we studied online pricing and allocation with perishable inventory and uniform price decisions, downstream transportation fulfillment, unknown linear demand, and uncensored observations. OCSAA combines counterfactual demand translation, ridge confidence, and a lower-confidence SAA objective to learn over the resulting non-convex stochastic program. The analysis establishes high-probability $\widetilde O(\sqrt T)$ regret, a polynomial-time additive implementation for rational polytopes, and minimax-optimal horizon dependence up to logarithmic factors.


\newpage
\bibliographystyle{splncs04}
\bibliography{ref}

\newpage
\appendix
\crefalias{section}{appendix}
\crefalias{subsection}{appendix}
\crefalias{subsubsection}{appendix}
{\huge\bfseries Appendix}
\addcontentsline{toc}{section}{Appendix}
\section{Proof Details}
\label{app:proofs}
This appendix gives the detailed arguments supporting the regret analysis in \Cref{sec:regret-analysis}.  The proof order follows the main-text roadmap: LP regularity, uniform SAA control, regression confidence, plug-in error, cumulative design uncertainty, and then the theorem, oracle, and lower-bound proofs.
\subsection{Proof of \Cref{lem:lp-regularity}}
\label{app:proof-lp-regularity}
First, for every feasible allocation $X$ in \eqref{eq:g-value},
\[
    \sum_{i=1}^m\sum_{j=1}^n X_{ij}
    \le \sum_{i=1}^m I_i
    \le \overline I_\Sigma.
\]
Since $\abs{C_{ij}-p}\le L_0$, every feasible allocation has objective value in $[-L_0\overline I_\Sigma,L_0\overline I_\Sigma]$.  The zero allocation is feasible, so the optimal value $G(\vI,p,\vD)$ lies in the same interval.  Adding $\abs{\ip{\vgamma}{\vI}}\le B_\gamma$ proves $\abs{\ell_{\vI,p}(\vN)}\le B_Q$.

We next prove Lipschitzness in demand.  Fix $p$ and $\vI$, and let $X$ be optimal for demand capacity $\vD$.  To make $X$ feasible for $\vD'$, remove arbitrary flow from any column $j$ whose total exceeds $D'_j$.  At most $(D_j-D'_j)_+$ units are removed from column $j$, and removing one unit can increase the objective by at most $L_0$.  Thus
\[
    g(\vI,p,\vD')
    \le g(\vI,p,\vD)+L_0\sum_{j=1}^n(D_j-D'_j)_+.
\]
Swapping $\vD$ and $\vD'$ gives
\[
    \abs{g(\vI,p,\vD)-g(\vI,p,\vD')}
    \le L_0\norm{\vD-\vD'}_1.
\]
The same argument applied to rows gives
\[
    \abs{g(\vI,p,\vD)-g(\vI',p,\vD)}
    \le L_0\norm{\vI-\vI'}_1.
\]
The triangle inequality proves the joint right-hand-side bound for non-negative demands.  Since the map $u\mapsto u^+$ is non-expansive in $\ell_1$, the demand part extends from $g$ to $G$ for arbitrary right-hand sides.

For price Lipschitzness with fixed non-negative $\vI,\vD$, the feasible set is unchanged.  Let $X_p$ be optimal at price $p$.  Then
\begin{align*}
    g(\vI,p',\vD)
    &\le \sum_{i,j}(C_{ij}-p')X_{p,ij}\\
    &=g(\vI,p,\vD)+(p-p')\sum_{i,j}X_{p,ij}\\
    &\le g(\vI,p,\vD)+\overline I_\Sigma\abs{p-p'}.
\end{align*}
Interchanging $p$ and $p'$ proves the price bound.

Finally, write $\Psi(z)=(\vI,p)$ and $\Psi(z')=(\vI',p')$.  The inventory cost and inventory right-hand-side terms contribute at most
\[
    \sum_{i=1}^m(\abs{\gamma_i}+L_0)\abs{I_i-I'_i}.
\]
For the price coordinate, first change the LP objective coefficient and then the demand vector.  By positive-part non-expansiveness and \Cref{ass:envelopes},
\[
    \norm{(\va^\star-\vb^\star p+\vN)^+
          -(\va^\star-\vb^\star p'+\vN)^+}_1
    \le \norm{\vb^\star}_1\abs{p-p'}
    \le B_b\abs{p-p'}.
\]
Thus
\[
    \abs{\ell_{\vI,p}(\vN)-\ell_{\vI',p'}(\vN)}
    \le \sum_{i=1}^m(\abs{\gamma_i}+L_0)\abs{I_i-I'_i}
      +(\overline I_\Sigma+L_0B_b)\abs{p-p'}.
\]
Substituting the normalized coordinates gives the claimed $L_{\rm dec}\norm{z-z'}_1$ bound.

\subsection{Proof of \Cref{lem:saa-main}}
\label{app:proof-uniform-saa}
The proof is carried out on the ambient box
\[
    \overline\cX:=\prod_{i=1}^m[0,\overline I_i]\times\cP,
\]
which contains the feasible set \(\cX=\cI\times\cP\).  The functions \(Q\) and
\(\widetilde Q_t\) are extended to \(\overline\cX\) by the same LP definitions
used in the main text.  After proving a uniform bound on \(\overline\cX\), we
restrict the result back to \(\cX\).

For $z\in[0,1]^d$, let
\[
    L(z):=Q(\Psi(z)),
    \qquad
    \widetilde L_t(z):=\widetilde Q_t(\Psi(z)).
\]
By \eqref{eq:oracle-demand-paper}, $\widetilde L_t(z)$ is the average of $t$
i.i.d. copies of the bounded loss $\ell_{\Psi(z)}(\vN)$, whose expectation is
$L(z)$.  At time $t$, define the deterministic grid
\[
    \cG_t:=\left\{0,\frac1t,\frac2t,\ldots,1\right\}^d,
    \qquad |\cG_t|=(t+1)^d.
\]
For any fixed grid point $g\in\cG_t$, Hoeffding's inequality for variables in
$[-B_Q,B_Q]$ gives
\[
    \Prob\left(
      \abs{\widetilde L_t(g)-L(g)}>
      B_Q\sqrt{\frac{2}{t}\log\frac{2(t+1)^d}{\delta_t^{\rm SAA}}}
    \right)
    \le \frac{\delta_t^{\rm SAA}}{(t+1)^d}.
\]
A union bound over $\cG_t$ shows that, with probability at least
$1-\delta_t^{\rm SAA}$, this bound holds for every grid point.

For arbitrary $z\in[0,1]^d$, choose a nearest grid point $g_t(z)\in\cG_t$ in
each coordinate.  Then $\norm{z-g_t(z)}_1\le d/t$.  By
\Cref{lem:lp-regularity}, both the empirical average and the population
expectation on the ambient box change by at most $L_{\rm dec}d/t$ when $z$ is
replaced by $g_t(z)$.  Therefore, on the grid event,
\[
    \abs{\widetilde L_t(z)-L(z)}
    \le
    B_Q\sqrt{\frac{2}{t}\log\frac{2(t+1)^d}{\delta_t^{\rm SAA}}}
    +\frac{2L_{\rm dec}d}{t}
    =\varepsilon_t^{\rm SAA}.
\]
Since the bound holds on the ambient box, it holds in particular on
$\cX=\cI\times\cP$.  Finally,
$\sum_{t\ge1}\delta_t^{\rm SAA}=\sum_{t\ge1}3\delta/(\pi^2t^2)=\delta/2$, so a
union bound over all $t$ proves the result.

\subsection{Auxiliary self-normalized inequality}
\label{app:proof-self-normalized}
We use the following standard finite-dimensional form, and include the proof for completeness.  Let $\eta_t$ be conditionally $\sigma$-sub-Gaussian given $\cF_{t-1}$, and let $x_t\in\R^r$ be $\cF_{t-1}$-measurable.  Define
\[
    S_t:=\sum_{s=1}^t x_s\eta_s,
    \qquad
    V_t:=\lambda I_r+\sum_{s=1}^t x_sx_s^\top.
\]
Then, for any $\delta_0\in(0,1)$, with probability at least $1-\delta_0$, simultaneously for all $t\ge0$,
\begin{equation}
    \norm{S_t}_{V_t^{-1}}^2
    \le
    2\sigma^2\log\left(
        \frac{\det(V_t)^{1/2}}{\det(\lambda I_r)^{1/2}\delta_0}
    \right).
    \label{eq:self-normalized-app}
\end{equation}

Fix $q\in\R^r$.  Conditional sub-Gaussianity implies that
\[
    M_t(q):=\exp\left(
       q^\top S_t-\frac{\sigma^2}{2}q^\top(V_t-\lambda I_r)q
    \right)
\]
is a non-negative supermartingale with initial value one.  Mix this supermartingale over the Gaussian density
\[
    f(q)=\left(\frac{\sigma^2\lambda}{2\pi}\right)^{r/2}
    \exp\left(-\frac{\sigma^2\lambda}{2}\norm{q}_2^2\right).
\]
By Tonelli's theorem, $\overline M_t:=\int M_t(q)f(q)dq$ is also a non-negative supermartingale.  Completing the square in the Gaussian integral yields
\[
    \overline M_t=
    \left(\frac{\det(\lambda I_r)}{\det(V_t)}\right)^{1/2}
    \exp\left(\frac{1}{2\sigma^2}\norm{S_t}_{V_t^{-1}}^2\right).
\]
Ville's maximal inequality gives $\Prob(\sup_t\overline M_t\ge1/\delta_0)\le\delta_0$.  On the complementary event, taking logarithms in the last display proves \eqref{eq:self-normalized-app}.  Bounded centered noise in \Cref{ass:noise} is conditionally sub-Gaussian by Hoeffding's lemma.

\subsection{Proof of \Cref{lem:ridge-main}}
\label{app:proof-ridge-confidence}
Fix demand class $j$.  Let
\[
    S_{t,j}:=\sum_{s=1}^t\phi(p_s)N_{s,j}.
\]
Using \eqref{eq:ridge-estimator-paper} and $Y_{s,j}=\phi(p_s)^\top\theta_j^\star+N_{s,j}$,
\[
    \widehat\theta_{t,j}^{\rm rid}-\theta_j^\star
    =V_t^{-1}S_{t,j}-\lambda V_t^{-1}\theta_j^\star.
\]
Therefore
\[
    \norm{\widehat\theta_{t,j}^{\rm rid}-\theta_j^\star}_{V_t}
    \le \norm{S_{t,j}}_{V_t^{-1}}
      +\lambda\norm{\theta_j^\star}_{V_t^{-1}}
    \le \norm{S_{t,j}}_{V_t^{-1}}+\sqrt\lambda S,
\]
where the last step uses $V_t\succeq\lambda I_2$ and \Cref{ass:envelopes}.  Apply the self-normalized inequality in \Cref{app:proof-self-normalized} with $x_s=\phi(p_s)$, $r=2$, and failure probability $\delta/(2n)$, then take a union bound over $j\in[n]$.  This proves the ridge-confidence statement with the definition of $\beta_t$ in \eqref{eq:beta-paper}.

For the unprojected ridge slope, note that
\[
    (\widehat b_{t,j}^{\rm rid}-b_j^\star)(p'-p)
    =\bigl(\phi(p)-\phi(p')\bigr)^\top
      (\widehat\theta_{t,j}^{\rm rid}-\theta_j^\star).
\]
Cauchy--Schwarz in the $V_t$ and $V_t^{-1}$ norms gives
\[
    \abs{(\widehat b_{t,j}^{\rm rid}-b_j^\star)(p'-p)}
    \le
    \norm{\widehat\theta_{t,j}^{\rm rid}-\theta_j^\star}_{V_t}
    \norm{\phi(p)-\phi(p')}_{V_t^{-1}}
    \le \beta_t\norm{\phi(p)-\phi(p')}_{V_t^{-1}}.
\]
Since $b_j^\star\in[0,\overline b_j]$ and Euclidean projection onto an interval is non-expansive,
\[
    |\widehat b_{t,j}-b_j^\star|
    =|\Pi_{[0,\overline b_j]}(\widehat b_{t,j}^{\rm rid})-\Pi_{[0,\overline b_j]}(b_j^\star)|
    \le |\widehat b_{t,j}^{\rm rid}-b_j^\star|.
\]
Multiplying by $|p'-p|$ proves the displayed slope consequence for the projected slope used by the algorithm.

\subsection{Proof of \Cref{lem:plugin-main}}
\label{app:proof-plugin-error}
By the right-hand-side Lipschitz part of \Cref{lem:lp-regularity}, for any fixed $t$, $\vI$, and $p$,
\begin{align*}
    \abs{\widehat Q_t(\vI,p)-\widetilde Q_t(\vI,p)}
    &\le \frac{L_0}{t}\sum_{s=1}^t
    \norm{\widehat\vD_{s,t}(p)-\vD_s^\star(p)}_1 \\
    &=\frac{L_0}{t}\sum_{s=1}^t\sum_{j=1}^n
    \abs{(\widehat b_{t,j}-b_j^\star)(p_s-p)}.
\end{align*}
On the event of \Cref{lem:ridge-main}, each summand is at most
\[
    \beta_t\norm{\phi(p)-\phi(p_s)}_{V_t^{-1}}.
\]
Hence
\begin{align*}
    \abs{\widehat Q_t(\vI,p)-\widetilde Q_t(\vI,p)}
    &\le L_0n\beta_t\cdot\frac1t\sum_{s=1}^t
    \norm{\phi(p)-\phi(p_s)}_{V_t^{-1}}\\
    &\le L_0n\beta_t
    \left[\frac1t\sum_{s=1}^t
    \norm{\phi(p)-\phi(p_s)}_{V_t^{-1}}^2\right]^{1/2},
\end{align*}
where the last step is Cauchy--Schwarz.  The final expression is $L_0n\beta_t\Gamma_t(p)$.

\subsection{Proof of \Cref{prop:confidence-band-main}}
\label{app:proof-confidence-band}
The event of \Cref{lem:saa-main} fails with probability at most $\delta/2$, and the event of \Cref{lem:ridge-main} fails with probability at most $\delta/2$.  On their intersection, for every $t$ and every $(\vI,p)$,
\[
    \abs{\widehat Q_t(\vI,p)-Q(\vI,p)}
    \le
    \abs{\widehat Q_t(\vI,p)-\widetilde Q_t(\vI,p)}
    +\abs{\widetilde Q_t(\vI,p)-Q(\vI,p)}.
\]
Apply \Cref{lem:plugin-main} to the first term and \Cref{lem:saa-main} to the second term to obtain
\[
    \abs{\widehat Q_t(\vI,p)-Q(\vI,p)}
    \le L_0n\beta_t\Gamma_t(p)+\varepsilon_t^{\rm SAA}=r_t(p).
\]
A union bound gives probability at least $1-\delta$.

\subsection{Proof of \Cref{lem:gamma-sum-main}}
\label{app:proof-design-sum}
For the action chosen after history $t$, Minkowski's inequality gives
\begin{align*}
    \Gamma_t(p_{t+1})
    &=\left[\frac1t\sum_{s=1}^t
      \norm{\phi(p_{t+1})-\phi(p_s)}_{V_t^{-1}}^2\right]^{1/2}\\
    &\le
      \norm{\phi(p_{t+1})}_{V_t^{-1}}
      +\left[\frac1t\sum_{s=1}^t
      \norm{\phi(p_s)}_{V_t^{-1}}^2\right]^{1/2}.
\end{align*}
For the historical term,
\begin{align*}
    \sum_{s=1}^t\norm{\phi(p_s)}_{V_t^{-1}}^2
    &=\tr\left(V_t^{-1}\sum_{s=1}^t\phi(p_s)\phi(p_s)^\top\right)\\
    &=\tr\left(V_t^{-1}(V_t-\lambda I_2)\right)
    \le 2.
\end{align*}
Thus its cumulative contribution is at most $\sum_{t=1}^{T-1}\sqrt{2/t}\le2\sqrt{2T}$.

For the current term, define
\[
    a_t^2:=\norm{\phi(p_{t+1})}_{V_t^{-1}}^2.
\]
Since $V_t\succeq\lambda I_2$, $a_t^2\le L_\phi^2/\lambda$.  The matrix determinant lemma yields
\[
    \log\frac{\det(V_T)}{\det(V_1)}
    =\sum_{t=1}^{T-1}\log(1+a_t^2).
\]
For $0\le x\le L_\phi^2/\lambda$,
\[
    \log(1+x)\ge \frac{x}{1+x}
    \ge \frac{x}{1+L_\phi^2/\lambda}.
\]
Therefore
\[
    \sum_{t=1}^{T-1}a_t^2
    \le \left(1+\frac{L_\phi^2}{\lambda}\right)
       \log\frac{\det(V_T)}{\det(V_1)}.
\]
Cauchy--Schwarz gives
\[
    \sum_{t=1}^{T-1}a_t
    \le
    \sqrt{T\left(1+\frac{L_\phi^2}{\lambda}\right)
       \log\frac{\det(V_T)}{\det(V_1)}}.
\]
It remains to upper bound the determinant ratio.  Let $\mu_1,\mu_2$ be the eigenvalues of
\[
    \sum_{s=1}^T\phi(p_s)\phi(p_s)^\top .
\]
Then $\mu_1+\mu_2\le TL_\phi^2$.  AM--GM gives
\[
    \frac{\det(V_T)}{\det(\lambda I_2)}
    =(1+\mu_1/\lambda)(1+\mu_2/\lambda)
    \le \left(1+\frac{TL_\phi^2}{2\lambda}\right)^2.
\]
Since $V_1\succeq\lambda I_2$, substituting this deterministic bound proves \eqref{eq:gamma-sum-main}.

\subsection{Proof of \Cref{thm:main-regret}}
\label{app:proof-main-theorem}

Define
\begin{align}
    A_T&:=\log\left(\frac{2\pi^2T^2(T+1)^d}{3\delta}\right),\nonumber\\
    \overline\beta_T&:=\sigma\sqrt{2\log\left(\frac{2n}{\delta}\left(1+\frac{TL_\phi^2}{2\lambda}\right)\right)}+\sqrt\lambda S,\nonumber\\
    H_T&:=\sqrt{2T\left(1+\frac{L_\phi^2}{\lambda}\right)\log\left(1+\frac{TL_\phi^2}{2\lambda}\right)}+2\sqrt{2T}.
    \label{eq:AT-beta-HT}
\end{align}
Work on the event of \Cref{prop:confidence-band-main}.  The first decision is arbitrary.  Since every population objective value lies in $[-B_Q,B_Q]$, the first-round regret is at most $2B_Q$.

For $t\ge1$, let $x_{t+1}:=(\vI_{t+1},p_{t+1})$ be the action selected by the LCB oracle after history $t$, and let $x^\star=(\vI^\star,p^\star)$.  The upper confidence bound at $x_{t+1}$ gives
\[
    Q(x_{t+1})\le\widehat Q_t(x_{t+1})+r_t(p_{t+1}).
\]
The optimality of the LCB oracle gives
\[
    \widehat Q_t(x_{t+1})-r_t(p_{t+1})
    \le \widehat Q_t(x^\star)-r_t(p^\star).
\]
The lower confidence bound at $x^\star$ gives
\[
    \widehat Q_t(x^\star)-r_t(p^\star)\le Q(x^\star).
\]
Combining the three inequalities yields the one-step regret bound
\begin{equation}
    Q(x_{t+1})-Q(x^\star)
    \le 2r_t(p_{t+1}).
    \label{eq:one-step-regret-app}
\end{equation}
Summing \eqref{eq:one-step-regret-app} gives
\begin{equation}
    R_T
    \le 2B_Q
      +2\sum_{t=1}^{T-1}\varepsilon_t^{\rm SAA}
      +2L_0n\sum_{t=1}^{T-1}\beta_t\Gamma_t(p_{t+1}).
    \label{eq:regret-before-sums-app}
\end{equation}
For $t\le T$, the logarithm in \eqref{eq:saa-radius-paper} is at most $A_T$.  Hence
\begin{align*}
    2\sum_{t=1}^{T-1}\varepsilon_t^{\rm SAA}
    &\le
    2\sum_{t=1}^{T-1}
    \left(B_Q\sqrt{\frac{2A_T}{t}}+\frac{2L_{\rm dec}d}{t}\right)\\
    &\le 4B_Q\sqrt{2TA_T}+4L_{\rm dec}d(1+\log T),
\end{align*}
using $\sum_{t=1}^{T-1}t^{-1/2}\le2\sqrt T$ and $\sum_{t=1}^{T-1}t^{-1}\le1+\log T$.

The determinant bound used in \Cref{app:proof-design-sum} implies $\beta_t\le\overline\beta_T$ for all $t\le T$.  Therefore, by \Cref{lem:gamma-sum-main},
\[
    \sum_{t=1}^{T-1}\beta_t\Gamma_t(p_{t+1})
    \le \overline\beta_T H_T.
\]
Substituting the last two bounds into \eqref{eq:regret-before-sums-app} gives the explicit bound
\begin{equation}
    R_T\le
    2B_Q+4B_Q\sqrt{2TA_T}+4L_{\rm dec}d(1+\log T)+2L_0n\overline\beta_T H_T.
    \label{eq:main-regret-bound}
\end{equation}
The order statement follows because $A_T=O(\log T)$, $\overline\beta_T=O(\sqrt{\log T})$, and $H_T=O(\sqrt{T\log T})$ for fixed model constants.

\subsection{Proof of \Cref{cor:approx-oracle}}
\label{app:proof-approx-oracle}
If the LCB minimization step is additively approximate, the LCB optimality inequality in the proof of \Cref{thm:main-regret} becomes
\[
    \widehat Q_t(x_{t+1})-r_t(p_{t+1})
    \le \widehat Q_t(x^\star)-r_t(p^\star)+\xi_t.
\]
The same confidence-band argument then gives
\[
    Q(x_{t+1})-Q(x^\star)
    \le 2r_t(p_{t+1})+\xi_t.
\]
Summing this inequality over $t=1,\ldots,T-1$ adds $\sum_{t=1}^{T-1}\xi_t$ to the regret bound.

\subsection{Proof of \Cref{prop:poly-additive-oracle}}
\label{app:proof-polynomial-oracle}
We prove the result in four steps.  Throughout the proof the history through
round $t$ is fixed, so $V_t$, $\beta_t$, $\Gamma_t$, and the projected slope
vector $\widehat\vb_t$ are deterministic quantities.

\paragraph{Fixed-price LP.}
Fix a price $q\in\cP$ and define
\[
    d_{s,j}(q):=[Y_{s,j}+\widehat b_{t,j}(p_s-q)]_+.
\]
For this fixed price, the clipped quantities $d_{s,j}(q)$ are constants.  By
the definition of $G$, the value
\[
    \min_{\vI\in\cI}
    \left\{
        \ip{\vgamma}{\vI}
        +\frac1t\sum_{s=1}^tG(\vI,q,\widehat\vD_{s,t}(q))
    \right\}
\]
is exactly the optimum of the LP in \Cref{alg:additive-lcb-oracle}: for each
historical sample $s$, the variables $X^s$ represent the allocation chosen in
the transportation problem defining $G(\vI,q,\widehat\vD_{s,t}(q))$, and all
samples share the same inventory vector $\vI$.  The price-dependent confidence
term $-L_0n\beta_t\Gamma_t(q)$ is constant with respect to the LP variables and
is added after the LP is solved.  If $\cI$ has a polynomial-size rational linear
description, the LP has $m+tmn$ variables, explicit row and column constraints
of order $t(m+n)$, and the constraints describing $\cI$.  Hence it is a
polynomial-size LP.

\paragraph{Price Lipschitzness of the fixed-price value.}
Let
\[
    H_t(p):=\min_{\vI\in\cI}\widehat Q_t(\vI,p)
\]
be the plug-in SAA value before subtracting the radius.  We first show that
$H_t$ is $(\overline I_\Sigma+L_0B_b)$-Lipschitz on $\cP$.  Fix $p,p'\in\cP$
and $\vI\in\cI$.  For a sample $s$, write
$\widehat\vD_s(p)=\vY_s+\widehat\vb_t(p_s-p)$.  By adding and subtracting
$G(\vI,p',\widehat\vD_s(p))$ and using \Cref{lem:lp-regularity},
\begin{align*}
    &\abs{G(\vI,p,\widehat\vD_s(p))-G(\vI,p',\widehat\vD_s(p'))} \\
    &\qquad\le
    \overline I_\Sigma|p-p'|+L_0\norm{\widehat\vD_s(p)^+-\widehat\vD_s(p')^+}_1.
\end{align*}
The positive-part map is non-expansive.  Moreover, the slope projection gives
$0\le\widehat b_{t,j}\le\overline b_j$ and hence
$\|\widehat\vb_t\|_1\le B_b$.  Therefore
\[
    \norm{\widehat\vD_s(p)^+-\widehat\vD_s(p')^+}_1
    \le B_b|p-p'|.
\]
Averaging over $s$ and adding the unchanged inventory cost gives, uniformly in
$\vI$,
\[
    |\widehat Q_t(\vI,p)-\widehat Q_t(\vI,p')|
    \le (\overline I_\Sigma+L_0B_b)|p-p'|.
\]
Taking minima over $\vI$ on the two sides preserves the same Lipschitz bound for
$H_t$.

It remains to control the price-dependent radius.  By the reverse triangle
inequality and the definition of $\Gamma_t$,
\begin{align*}
    |\Gamma_t(p)-\Gamma_t(p')|
    &\le \left[
        \frac1t\sum_{s=1}^t
        \norm{\phi(p)-\phi(p')}_{V_t^{-1}}^2
    \right]^{1/2} \\
    &=\norm{\phi(p)-\phi(p')}_{V_t^{-1}}
    \le \frac{|p-p'|}{\sqrt\lambda},
\end{align*}
because $V_t\succeq\lambda I_2$ and
$\phi(p)-\phi(p')=(0,p'-p)^\top$.  The SAA term $\varepsilon_t^{\rm SAA}$ is
action-independent.  Therefore the LCB profile
\[
    F_t(p):=\min_{\vI\in\cI}\{\widehat Q_t(\vI,p)-r_t(p)\}
    =H_t(p)-\varepsilon_t^{\rm SAA}-L_0n\beta_t\Gamma_t(p)
\]
is $L_t^{\rm price}$-Lipschitz, with $L_t^{\rm price}$ defined in
\eqref{eq:price-lipschitz-constant}.  In particular, $F_t$ is continuous on the
compact interval $\cP$, so a minimizer $p^\star\in\argmin_{p\in\cP}F_t(p)$
exists.

\paragraph{Discretization error.}
Let $q\in\cP_t^\xi$ be a grid point within distance at most $h_t$ of
$p^\star$.  Since the grid contains the endpoints and has mesh at most $h_t$,
such a point exists.  By Lipschitzness,
\[
    F_t(q)\le F_t(p^\star)+L_t^{\rm price}h_t
    \le F_t(p^\star)+\xi_t/2.
\]
The algorithm chooses the best grid price, so its returned LCB value is no
larger than $F_t(q)$.  Hence the returned action is in fact $\xi_t/2$-optimal,
and therefore $\xi_t$-optimal.  Omitting $\varepsilon_t^{\rm SAA}$ inside
\Cref{alg:additive-lcb-oracle} does not change value gaps because that term is
action-independent.

\paragraph{Complexity.}
A uniform grid of mesh $h_t$ over an interval of length
$\Delta_p=\overline p-\underline p$ has at most
\[
    2+\left\lceil \frac{\Delta_p}{h_t}\right\rceil
    =O\!\left(1+\frac{\Delta_p L_t^{\rm price}}{\xi_t}\right)
\]
points.  At each point the oracle solves one polynomial-size LP, as shown in the
fixed-price step.  This proves the complexity claim.

\begin{remark}[Arithmetic model and numerical LP accuracy]
The proposition's complexity statement is in the real-arithmetic model; for
rational inputs of polynomial encoding length, it is the usual polynomial LP
solvability statement.  With only a separation oracle for $\cI$, the same
fixed-price linear optimization problem can be solved by the ellipsoid method
under standard bit-complexity assumptions.  If the fixed-price LPs are solved
only to numerical accuracy $\eta_t$, the same proof gives a $\xi_t+\eta_t$
additive oracle; one can split the accuracy budget by using mesh
$\xi_t/(4L_t^{\rm price})$ and LP accuracy $\xi_t/2$.
\end{remark}

\subsection{Proof of \Cref{thm:lower-bound}}
\label{app:proof-lower-bound}

We first prove the scalar continuous-inventory lower bound and then give the
extension to general dimensions.  Both constructions use uncensored demand observations and keep the demand slope identical across the hard instances.  The common slope is strictly positive and lies in the known envelope, but it may depend on the horizon and decrease with $T$.

\subsubsection*{Scalar continuous-inventory construction.}
Fix a horizon $T$ and set
\[
    \varepsilon:=\frac{1}{8\sqrt T},
    \qquad
    q_+:=\frac12+\varepsilon,
    \qquad
    q_-:=\frac12-\varepsilon,
    \qquad
    b_T:=\frac{\varepsilon^2}{128}.
\]
For all sufficiently large $T$, $\varepsilon\le 1/4$ and $0<b_T\le1$.  The
model has $m=n=1$, continuous inventory set $\cI=[0,1]$, price interval
$\cP=[0,1]$, allocation cost $C=0$, and inventory cost $\gamma=1/2$.  Under
instance $\nu_\pm$, let $Z_t\sim\mathrm{Bernoulli}(q_\pm)$ independently over
time and define
\[
    N_t:=Z_t-q_\pm,
    \qquad
    a_\pm:=q_\pm+b_T,
    \qquad
    b^\star:=b_T .
\]
Then $\E_{\nu_\pm}[N_t]=0$ and $|N_t|\le1$.  For every price $p\in[0,1]$,
\[
    D_t(p)=a_\pm-b_Tp+N_t=Z_t+b_T(1-p)\ge0.
\]
Thus the two instances satisfy bounded mean-zero noise, non-negative true demand,
positive slope, and uncensored feedback.  The slope is the same in the two instances; the lower bound is therefore not driven by slope learning.

For $I\in[0,1]$, $p\in[0,1]$, and $h=b_T(1-p)$, the scalar allocation value is
$g(I,p,D)=-p\min\{I,D\}$.  Hence, writing $q$ for either $q_+$ or $q_-$,
\begin{equation}
    Q_q(I,p)
    =\frac12 I
     -p\Bigl(qI+(1-q)\min\{I,b_T(1-p)\}\Bigr).
    \label{eq:lb-cont-Q}
\end{equation}
Indeed, if $Z=1$, then $Z+h\ge1$ and $\min\{I,Z+h\}=I$; if $Z=0$, the clipped
quantity is $\min\{I,h\}$.  Let
\[
    Q_q^0(I,p):=\left(\frac12-pq\right)I
\]
be the value obtained from \eqref{eq:lb-cont-Q} when $b_T=0$.  Since the extra
term in \eqref{eq:lb-cont-Q} is non-negative and at most $b_T$, uniformly over
$I,p,q$,
\begin{equation}
    Q_q^0(I,p)-b_T\le Q_q(I,p)\le Q_q^0(I,p).
    \label{eq:lb-cont-perturb}
\end{equation}

Under $q_+$, the action $(I,p)=(1,1)$ has value
$Q_{q_+}(1,1)=1/2-q_+=-\varepsilon$, so the optimal value is at most
$-\varepsilon$.  Moreover, for every $I,p$,
\[
    Q_{q_+}^0(I,p)
    =\left(\frac12-p\left(\frac12+\varepsilon\right)\right)I
    \ge -\varepsilon I.
\]
Combining this with \eqref{eq:lb-cont-perturb}, the instantaneous regret under
$\nu_+$ satisfies
\begin{equation}
    Q_{q_+}(I,p)-\min_{I',p'}Q_{q_+}(I',p')
    \ge \varepsilon(1-I)-b_T .
    \label{eq:lb-cont-plus-reg}
\end{equation}
Under $q_-$, the zero-inventory action has value zero, so the optimal value is
at most zero.  Also,
\[
    Q_{q_-}^0(I,p)
    =\left(\frac12-p\left(\frac12-\varepsilon\right)\right)I
    \ge \varepsilon I.
\]
Thus
\begin{equation}
    Q_{q_-}(I,p)-\min_{I',p'}Q_{q_-}(I',p')
    \ge Q_{q_-}(I,p)
    \ge \varepsilon I-b_T .
    \label{eq:lb-cont-minus-reg}
\end{equation}
The continuous inventory variable therefore plays the same role as a randomized
or fractional decision between the two inventory regimes.

Let $I_t\in[0,1]$ be the inventory chosen by an arbitrary non-anticipatory policy
in round $t$.  From \eqref{eq:lb-cont-plus-reg}--\eqref{eq:lb-cont-minus-reg},
\begin{equation}
\begin{aligned}
    \E_+[R_T]+\E_-[R_T]
    &\ge
    \varepsilon\sum_{t=1}^T
    \left(\E_+[1-I_t]+\E_-[I_t]\right)-2b_TT .
\end{aligned}
\label{eq:lb-cont-reg-sum}
\end{equation}
Let $\mathbb P_\pm^{t-1}$ be the law of the history available before choosing
the round-$t$ action.  For any $[0,1]$-valued history-measurable random variable
$A_t$,
\[
    \E_+[1-A_t]+\E_-[A_t]
    =1-\E_+[A_t]+\E_-[A_t]
    \ge 1-\mathrm{TV}(\mathbb P_+^{t-1},\mathbb P_-^{t-1}).
\]
Applying this with $A_t=I_t$, we need only control the total variation distance
between the two adaptive history laws.

Conditional on the past and on the chosen price $p_s$, the uncensored
observation is
\[
    Y_s=Z_s+b_T(1-p_s).
\]
The shift $b_T(1-p_s)$ is deterministic and identical under the two instances,
so the conditional KL divergence is exactly
$\mathsf{kl}(q_+,q_-)$.  By the adaptive chain rule,
\[
    \mathrm{KL}(\mathbb P_+^{t-1}\|\mathbb P_-^{t-1})
    \le (t-1)\mathsf{kl}(q_+,q_-).
\]
For $\varepsilon\le1/4$,
\[
    \mathsf{kl}\left(\frac12+\varepsilon,\frac12-\varepsilon\right)
    =2\varepsilon\log\frac{1+2\varepsilon}{1-2\varepsilon}
    \le16\varepsilon^2.
\]
Therefore, for all $t\le T$,
\[
    \mathrm{KL}(\mathbb P_+^{t-1}\|\mathbb P_-^{t-1})
    \le16T\varepsilon^2=\frac14.
\]
Pinsker's inequality gives
$\mathrm{TV}(\mathbb P_+^{t-1},\mathbb P_-^{t-1})\le\sqrt{1/8}<1/2$.
Substituting into \eqref{eq:lb-cont-reg-sum},
\[
    \E_+[R_T]+\E_-[R_T]
    \ge \frac{\varepsilon T}{2}-2b_TT .
\]
Since $\varepsilon=1/(8\sqrt T)$ and $b_TT=\varepsilon^2T/128=1/8192$, at
least one of the two instances satisfies, for all sufficiently large $T$,
\[
    \max\{\E_+[R_T],\E_-[R_T]\}
    \ge \frac12\left(\frac{\varepsilon T}{2}-2b_TT\right)
    \ge c\sqrt T
\]
for a universal numerical constant $c>0$, for instance any small enough
$c\le1/128$.  This proves the scalar continuous-inventory part of the theorem.

\subsubsection*{Dimension-dependent extension.}
Let $K:=\min\{m,n\}$.  We construct a $2^K$-instance subclass by placing $K$
independent copies of the scalar hard problem on diagonal resource--class pairs.
The feasible set is $\cI=[0,1]^m$ and $\cP=[0,1]$.  For active resources
$i\le K$, set $\gamma_i=1/2$; for inactive resources $i>K$, set
$\gamma_i=1$.  Set diagonal active costs $C_{jj}=0$ for $j\le K$ and set every
other cost to $2$, which is larger than all feasible prices.  Hence all
non-diagonal and inactive arcs are never used by the transportation LP, because
the zero flow on those arcs is feasible and their objective coefficients are
strictly positive.

For each sign vector $\omega\in\{+1,-1\}^K$, define
$q_{\omega_j}=1/2+\omega_j\varepsilon$ for active class $j\le K$.  Active-class
noises are independent over classes and time, with
$Z_{t,j}\sim\mathrm{Bernoulli}(q_{\omega_j})$ and
$N_{t,j}=Z_{t,j}-q_{\omega_j}$.  Set $a_j=q_{\omega_j}+b_T$ and
$b_j^\star=b_T$ for $j\le K$.  Extra demand classes, if any, may be assigned bounded, mean-zero noise, intercept $1+b_T$, and slope $b_T$; they
are economically inactive because all arcs to them have cost $2$.  All slopes
are therefore positive and identical across the family.

For each instance $\omega$, the population loss decomposes over the active
diagonal pairs, up to non-negative inventory costs on inactive resources.  The
scalar bounds \eqref{eq:lb-cont-plus-reg}--\eqref{eq:lb-cont-minus-reg} imply
that in every round,
\begin{equation}
    r_t(\omega)
    \ge
    \varepsilon\sum_{j:\omega_j=+1}(1-I_{t,j})
    +\varepsilon\sum_{j:\omega_j=-1}I_{t,j}
    -Kb_T .
    \label{eq:lb-hypercube-instant}
\end{equation}
Average \eqref{eq:lb-hypercube-instant} over the $2^K$ instances.  For a sign
vector $\omega$ with $\omega_j=+1$, let $\omega^{(j)}$ be the same vector with
coordinate $j$ flipped.  As before, for every $t$,
\[
    \E_\omega[1-I_{t,j}]+\E_{\omega^{(j)}}[I_{t,j}]
    \ge
    1-\mathrm{TV}(\mathbb P_\omega^{t-1},\mathbb P_{\omega^{(j)}}^{t-1}).
\]
The two histories differ only in the $j$th active demand coordinate.  Conditional
on any adaptive price, that coordinate is again a common deterministic shift of
a Bernoulli draw, so
\[
    \mathrm{KL}(\mathbb P_\omega^{t-1}\|\mathbb P_{\omega^{(j)}}^{t-1})
    \le (t-1)\mathsf{kl}(q_+,q_-)
    \le\frac14,
\]
and the corresponding total variation distance is less than $1/2$.  Hence the
average regret over the hypercube is at least
\[
    \frac{1}{2^K}\sum_{\omega\in\{\pm1\}^K}\E_\omega[R_T]
    \ge
    T\left(\frac{K\varepsilon}{4}-Kb_T\right)
    \ge c'K\sqrt T
\]
for a universal constant $c'>0$ and all sufficiently large $T$.  Therefore some
instance in the family has expected regret at least $c'K\sqrt T$.  Since
$K=\min\{m,n\}$, this gives the claimed
$\Omega(\min\{m,n\}\sqrt T)$ dimension-dependent lower bound.

\newpage
\section{Additional Numerical Details}
\label{app:numerical}
This appendix collects numerical summaries supporting the diagnostic discussion in the main text.  None of these tables is used in the theoretical results; they are included to make the plotted finite-horizon diagnostics easier to audit.
\subsection{Benchmark and implementation details}
\label[appendix]{app:reproducibility}
This subsection records the numerical configuration used for the diagnostic plots and tables.  The plotting pass used saved trajectories only, with matched random seeds across policies; no new simulations are needed to reproduce the figures from the saved trajectory files.

\paragraph{Scalar benchmark.}
The scalar benchmark has one supply node and one demand class, with
\[
    m=n=1,\qquad \cI=[0,8],\qquad \cP=[0.5,6.0],\qquad
    \gamma=0.8,\qquad C=1.0.
\]
The true demand parameters are
\[
    a^\star=10.0,\qquad b^\star=1.2,
\]
with known slope envelope \(\overline b=2.0\) and parameter-norm bound \(S=\sqrt{12^2+2^2}=\sqrt{148}\).  The noise is continuous uniform, \(N_t\sim {\rm Unif}[-0.8,0.8]\), so the simulation uses closed-form population evaluation rather than a finite noise support.  The minimum true demand over the price interval and noise support is \(10-1.2\cdot 6-0.8=2.0\), so true demand is always positive.  The initial action is \((I_1,p_1)=(4.0,3.25)\).  The allocation and empirical objectives use the positive-part convention for translated demands, and the scalar implementation also clips inventory at zero in numerical routines.

\paragraph{Two-by-two benchmark.}
The two-by-two benchmark has two supply nodes and two demand classes.  The implemented inventory search is over the rectangular grid in \([0,9]^2\), with no coupling constraint across resources.  The price and inventory grids are
\[
\begin{aligned}
    \cP_{\rm grid} &= \{3.5,3.875,4.25,4.625,5.0,5.375,5.75,6.125,6.5,6.875,7.25,7.625,8.0\},\\
    \cI_{\rm grid} &= \{0,1.5,3.0,4.5,6.0,7.5,9.0\}^2 .
\end{aligned}
\]
The cost and demand parameters are
\[
    \vgamma=(0.30,0.35),\qquad
    C=\begin{pmatrix}2.0&3.2\\ 3.1&2.1\end{pmatrix},\qquad
    \va^\star=(10.0,9.0),\qquad
    \vb^\star=(1.0,0.8).
\]
The implementation uses a parameter bound \(\theta_{\rm bd}=12.0\) for confidence parameters rather than a separately named two-by-two slope-envelope field.  Noise coordinates are independent, and each coordinate has support \(\{-0.25,0,0.25\}\) with probabilities \((0.25,0.50,0.25)\), giving nine joint scenarios.  The allocation LP uses \([D_j]_+\) in every scenario.  Demand is always positive on the price interval and noise support, with coordinatewise minima \(1.75\) and \(2.35\).  The initial action is \((\vI_1,p_1)=((4.5,4.5),5.75)\).

\paragraph{Evaluation grid and pseudo-regret.}
In the scalar benchmark, the evaluation grid uses \(41\) equally spaced prices in \([0.5,6.0]\); for each grid price, the inventory component is optimized exactly over \([0,8]\) under the closed-form population objective.  Thus the scalar grid benchmark is the minimum over a finite price grid with exact scalar population-optimal inventory at each grid price.  In the two-by-two benchmark, the benchmark is the minimum over the \(13\times 7\times 7\) finite action grid above, and \(Q(\vI,p)\) is evaluated exactly over the nine finite noise scenarios.  Both benchmarks use the same grid pseudo-regret convention
\[
    R_t^{\pi,{\rm grid}}
    = \sum_{s=1}^{t} \max\{0,\, Q(\vI_s^\pi,p_s^\pi)-Q_{\rm grid}^\star\},
    \qquad
    Q_{\rm grid}^\star := \min_{(\vI,p)\in \cA_{\rm grid}} Q(\vI,p).
\]
The implementation checks that any negative numerical gap is no smaller than \(-10^{-8}\).  Cumulative regret is computed for each seed trajectory first and then averaged across seeds at each checkpoint.

\paragraph{Horizons, checkpoints, and seeds.}
The scalar main trajectories come from restarted horizons \(256,512,1024,2048\), and the final figure uses the longest saved trajectory, \(T_{\max}=2048\), for each policy and seed.  The scalar seeds are \(22345,22346,22347,22348,22349,22350\), and the saved scalar trajectories contain every round.  The two-by-two main run uses horizon \(T_{\max}=1024\) with seeds \(32345,32346,32347\) and checkpoints
\[
    1,2,3,4,6,8,12,16,24,32,48,64,96,128,192,256,384,512,768,1024.
\]
Seeds are matched across policies, and the same demand-noise sequence is used across policies for a fixed seed.  The scalar endpoint summary in \Cref{tab:endpoint-summary} uses restarted horizons \(256,512,1024,2048\).  The saved scalar mechanism run used horizons \(512,1024,2048,4096\) and seeds \(22345,22346,22347,22348\).

\paragraph{Policies and hyperparameters.}
OCSAA is implemented as the lower-confidence counterfactual SAA policy.  Both benchmarks use ridge regularization \(\lambda=1.0\) and confidence level \(\delta=0.05\).  The scalar run uses \(\sigma=0.8\), \(S=\sqrt{148}\), and \(L_0=5.0\).  The two-by-two run uses \(\sigma=0.25\), \(S=\sqrt{2}\cdot 12\), and \(L_0=6.0\), with an \(n=2\) multiplier in the parameter-radius term.  The action-independent SAA radius is omitted from the argmin, as in the theory, because it does not affect comparisons across actions.  The final grid-based diagnostics do not use a separate recorded \(\xi_t\) schedule: the scalar policy searches the stated scalar price grid with the scalar inventory shortcut, and the two-by-two policy searches the stated finite action grid.  Slope projection was not activated in the saved main policies; the reported simulations use the ridge slope estimates directly in counterfactual translation.

\textsc{Greedy-SAA} minimizes the same counterfactual plug-in SAA objective without the lower-confidence radius, using ridge-estimated slopes for translation.  In the scalar implementation, ties are broken by the first encountered grid price and by the inventory-oracle candidate order consisting of zero inventory, maximum inventory, and an empirical quantile candidate.  In the two-by-two implementation, ties are broken lexicographically by the tuple consisting of objective value, price, inventory coordinates, and action-grid index.  \textsc{Oracle-slope-SAA} uses the true slope vector in the counterfactual translation and has no active lower-confidence term in the final implementation.

\paragraph{Optimization and numerical tolerances.}
The scalar runs do not call an LP solver for the main evaluation; they use scalar formulas and an exact empirical inventory shortcut.  A separate continuous diagnostic uses \texttt{scipy.optimize.minimize\_scalar} with method \texttt{bounded} and \texttt{xatol=1e-12}.  In the two-by-two runs, allocation values are computed by exact vertex enumeration of the \(2\times2\) allocation LP, with feasibility tolerance \(10^{-9}\).  A SciPy \texttt{linprog} helper with method \texttt{highs} exists in the code path, but no explicit HiGHS tolerances or time limits are set; a solver failure raises a runtime error.  The archived final-run manifests do not record the original package versions.  The inspected plotting environment used NumPy 1.26.4, SciPy 1.13.1, pandas 2.2.2, and matplotlib 3.9.2.

\subsection{Tail-fit summaries}
\label[appendix]{app:tailfit-summary}
The displayed checkpoints follow the pre-specified schedule
\[
    1,2,3,4,6,8,12,16,24,32,48,64,96,128,192,256,384,512,768,1024,1536,2048,
\]
with the first ten checkpoints dropped for both benchmarks.  Thus the scalar displayed and fitted checkpoints are \(48,64,96,128,192,256,384,512,768,1024,1536,2048\), and the two-by-two displayed and fitted checkpoints are \(48,64,96,128,192,256,384,512,768,1024\).  Dotted fits use exactly these displayed checkpoints.  The table below reports finite-window descriptive slopes from fitting \(\log \overline R_t\) on \(\log t\), together with seed-bootstrap intervals computed from 1000 resamples using bootstrap seed 20260629.
\begin{table}[H]
    \centering
    \caption{Displayed-window slope summaries for the scalar and two-by-two benchmarks.  The slopes are finite-window descriptive quantities and should not be interpreted as asymptotic rate estimates.}
    \label{tab:experiment-fit-summary}
    \resizebox{0.82\linewidth}{!}{%
    \begin{tabular}{llcc}
        \toprule
        Benchmark & Policy & Displayed checkpoints & Slope and bootstrap interval \\
        \midrule
        Scalar & OCSAA & $48$--$2048$ & $0.488\;[0.484,0.491]$ \\
        Scalar & \textsc{Greedy-SAA} & $48$--$2048$ & $0.909\;[0.903,0.914]$ \\
        Scalar & \textsc{Oracle-slope-SAA} & $48$--$2048$ & $0.083\;[0.044,0.132]$ \\
        Two-by-two & OCSAA & $48$--$1024$ & $0.545\;[0.541,0.548]$ \\
        Two-by-two & \textsc{Greedy-SAA} & $48$--$1024$ & $0.977\;[0.976,0.977]$ \\
        Two-by-two & \textsc{Oracle-slope-SAA} & $48$--$1024$ & $0.000\;[0.000,0.000]$ \\
        \bottomrule
    \end{tabular}}
\end{table}

\subsection{Endpoint summaries}
\label[appendix]{app:endpoint-mechanism-figures}
The endpoint summary is a separate scalar diagnostic based on restarted horizon runs.  It is included only as a finite-horizon consistency check for the behavior seen in the checkpoint trajectories.
\begin{table}[H]
    \centering
    \caption{Scalar restarted endpoint summary.  Fitted exponents are descriptive finite-horizon summaries from restarted endpoint horizons, not asymptotic estimates.}
    \label{tab:endpoint-summary}
    \resizebox{0.88\linewidth}{!}{\begin{tabular}{lrrrrr}
\toprule
Algorithm & $H=256$ & $H=512$ & $H=1024$ & $H=2048$ & Fitted exponent \\
\midrule
OCSAA & 410.7 & 603.3 & 898.1 & 1,347.0 & 0.571 \\
\textsc{Greedy-SAA} & 299.0 & 573.2 & 1,121.1 & 2,215.9 & 0.964 \\
\textsc{Oracle-slope-SAA} & 11.3 & 12.3 & 12.9 & 13.4 & 0.078 \\
\bottomrule
\end{tabular}
}
\end{table}

\subsection{Normalized regret diagnostics}
\begin{figure}[H]
    \centering
    \includegraphics[width=0.98\linewidth]{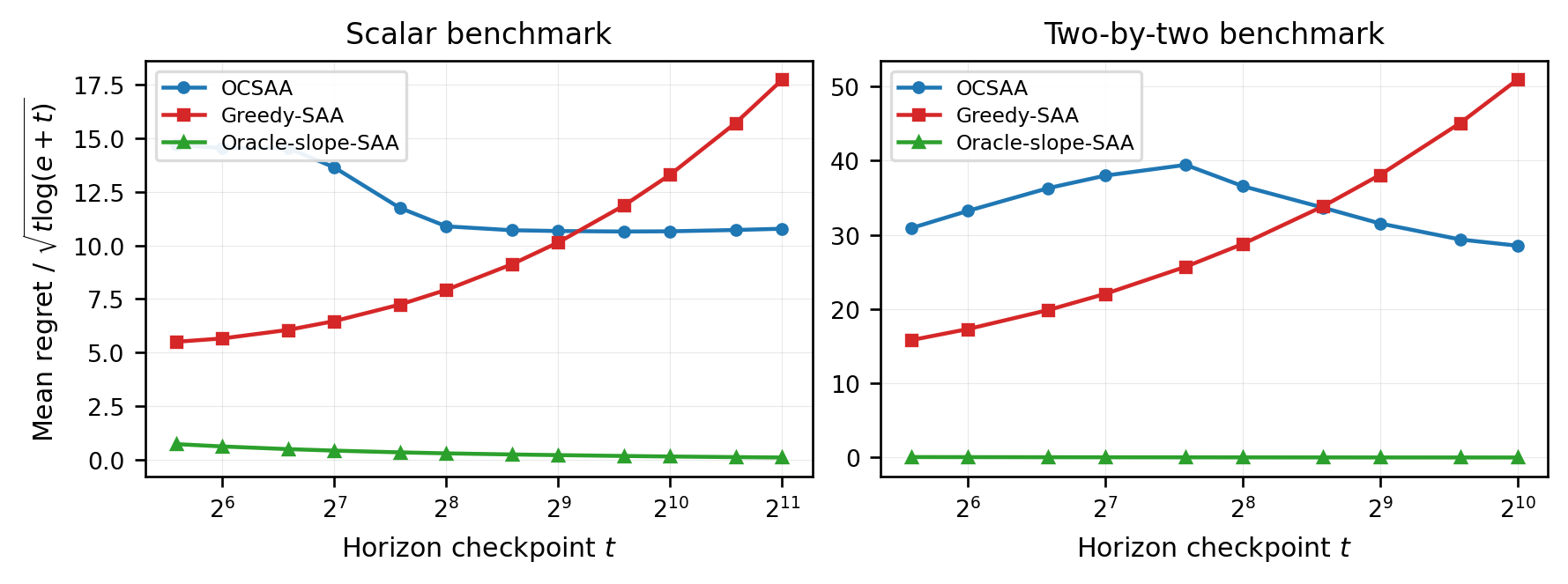}
    \caption{Normalized grid pseudo-regret diagnostics for the scalar benchmark and the two-by-two benchmark.  The normalization is \(\sqrt{t\log(e+t)}\), matching the theorem-scale dependence up to logarithmic factors.  These plots are diagnostic and should not be read as asymptotic empirical rate estimates.}
    \label{fig:normalized-regret}
\end{figure}

\label[appendix]{app:normalized-regret}
\Cref{fig:normalized-regret} plots mean grid pseudo-regret normalized by \(\sqrt{t\log(e+t)}\), matching the theorem scale up to logarithmic factors.  The curves should be read as stability diagnostics on the displayed horizons, not as asymptotic rate evidence.

\subsection{Mechanism ablations}
\label[appendix]{app:mechanism-tables}
The mechanism runs compare OCSAA with ablations that keep only selected components of the lower-confidence rule.  The parameter-radius-only ablation uses the same action rule as OCSAA in the saved runs; the SAA-radius-only ablation has the same behavior as \textsc{Greedy-SAA} because the SAA term is action-independent; and the oracle-slope LCB ablation has the same behavior as \textsc{Oracle-slope-SAA} because no active LCB radius is used for that final oracle-slope implementation.  The reported overlap is equality of seed-averaged checkpoint cumulative regret, not a separate action-level equivalence theorem.  The overlap tolerance is \(10^{-8}\); the saved two-by-two identity diagnostic used \(10^{-9}\).  All saved maximum gaps are zero.

\begin{table}[H]
    \centering
    \caption{Mechanism-ablation overlap checks.  ``Overlap'' means that the seed-averaged checkpoint regret trajectories coincide on the tested runs.}
    \label{tab:mechanism-summary}
    \resizebox{\linewidth}{!}{\begin{tabular}{lll}
\toprule
Benchmark & Pair & Empirical relation on tested runs \\
\midrule
Scalar & OCSAA vs. parameter-radius-only ablation & Overlap; max gap $0$ \\
Scalar & \textsc{Greedy-SAA} vs. SAA-radius-only ablation & Overlap; max gap $0$ \\
Scalar & \textsc{Oracle-slope-SAA} vs. oracle-slope LCB ablation & Overlap; max gap $0$ \\
Two-by-two & OCSAA vs. parameter-radius-only ablation & Overlap; max gap $0$ \\
Two-by-two & \textsc{Greedy-SAA} vs. SAA-radius-only ablation & Overlap; max gap $0$ \\
Two-by-two & \textsc{Oracle-slope-SAA} vs. oracle-slope LCB ablation & Overlap; max gap $0$ \\
\bottomrule
\end{tabular}
}
\end{table}

The radius diagnostics below provide a complementary check on the scale of the proof radii in the scalar runs.  They are reported only to document conservatism of the confidence terms on saved checkpoints and are not used to tune any policy.
\begin{table}[H]
    \centering
    \caption{Scalar radius diagnostics.  The ratios compare measured uniform errors with their proof radii over saved checkpoints.}
    \label{tab:radius-diagnostics}
    \begin{tabular}{lr}
\toprule
Diagnostic & Mean over checkpoints \\
\midrule
$E_{\rm SAA}/r_{\rm SAA}$ & 0.0066 \\
Parameter-radius ratio & 0.0772 \\
Total-radius ratio & 0.0166 \\
\bottomrule
\end{tabular}


\end{table}

\newpage
\section{More Discussions}
This appendix discusses related literature that is broadly relevant to the theme of this work, as well as current limitations and potential extensions, complementing \Cref{sec:related-work} and \Cref{sec:discussion}, respectively.
\subsection{Extended Related Work}
\label{app:extended-related-work}
Here we record additional papers that are relevant to the modeling ingredients in \Cref{sec:related-work}, but are less central than the closest comparisons in the main text.

\paragraph{Dynamic pricing with unknown demand.}
Kleinberg and Leighton study online posted-price learning and quantify the regret cost of not knowing the demand curve \cite{kleinberg2003value}.  Besbes and Zeevi develop risk bounds and near-optimal algorithms for pricing without knowing the demand function \cite{besbes2009dynamic}.  Broder and Rusmevichientong analyze dynamic pricing under a general parametric choice model \cite{broder2012dynamic}.  Keskin and Zeevi propose asymptotically optimal semi-myopic policies for unknown demand parameters \cite{keskin2014dynamic}.  Wang, Deng, and Ye design a learning-while-doing algorithm for single-product revenue management \cite{wang2014close}.  Wang, Chen, and Simchi-Levi study multimodal pricing, where the revenue objective can have multiple local optima \cite{wang2021multimodal}.  These papers motivate our demand-learning component, but they do not include a downstream transportation LP whose value changes with price.

\paragraph{Contextual and high-dimensional pricing.}
Cohen, Lobel, and Paes Leme introduce feature-based dynamic pricing \cite{cohen2020feature,cohen2020feature_journal}.  Javanmard and Nazerzadeh study high-dimensional dynamic pricing with structured parameters \cite{javanmard2019dynamic}, and Javanmard et al extends this line to multi-product pricing with heterogeneous sensitivities \cite{javanmard2020multi}. Xu and Wang further generalize the results on adversarial inputs \cite{xu2021logarithmic} \cite{xu2024pricing} and agnostic distributions \cite{xu2022towards} \cite{xu2026optimal}. Qiang and Bayati analyze pricing with demand covariates \cite{qiang2016dynamic}, while Wang, Talluri, and Li give regret bounds for covariate-based pricing without i.i.d. covariate arrivals \cite{wang2021dynamic}. Baby et al consider the non-stationarity in contextual pricing \cite{baby2023non}.  Ban and Keskin study personalized pricing with machine-learning features and heterogeneous elasticities \cite{ban2021personalized}.  Shah, Johari, and Blanchet propose semi-parametric contextual pricing \cite{shah2019semi}, and Bu, Simchi-Levi, and Wang study partially linear contextual demand \cite{bu2022context}.  These works exploit covariates or personalization, whereas our paper keeps one uniform price and focuses on non-convex recourse.

\paragraph{Pricing with inventory, capacity, and censoring.}
Gallego and van Ryzin analyze finite-horizon dynamic pricing of inventories under stochastic demand \cite{gallego1994optimal}, and Elmaghraby and Keskinocak survey pricing with inventory considerations \cite{elmaghraby2003dynamic}.  Chen and Gallego develop a primal--dual learning algorithm for personalized pricing with inventory constraints \cite{chen2018primal}.  Chen, Jasin, and Duenyas study multi-product pricing with finite resource capacity and unknown demand \cite{chen2021joint}, and their non-parametric work develops self-adjusting controls for finite-capacity multi-product pricing \cite{chen2019nonparametric}.  Chen, Chao, and Ahn coordinate pricing and inventory replenishment under non-parametric demand learning \cite{chen2019coordinating}. Besbes and Muharremoglu analyze demand censoring in the newsvendor problem \cite{besbes2013implications}, while Chen et al study pricing--inventory control under censored demand \cite{chen2020data,chen2023optimal} and Xu et al generalize this setting to arbitrarily non-stationary inventories \cite{xu2025dynamic}.  These models study inventory or lost-sales feedback, whereas our setting assumes per-period inventory and full uncensored demand.

\paragraph{Network revenue management and online allocation.}
Talluri and van Ryzin provide a comprehensive treatment of revenue-management models and controls \cite{talluri2006theory}.  Reiman and Wang propose an asymptotically optimal policy for quantity-based network revenue management \cite{reiman2008asymptotically}, and Jasin and Kumar analyze re-solving heuristics with bounded revenue loss \cite{jasin2012re}.  Bumpensanti and Wang show that thresholded re-solving can achieve uniformly bounded loss \cite{bumpensanti2020re}.  Ferreira, Simchi-Levi, and Wang apply Thompson sampling to online network revenue management \cite{ferreira2018online}.  Vera, Banerjee, and Gurvich use Bellman inequalities to obtain constant-regret online allocation and pricing policies \cite{vera2021online}.  Jiang, Ma, and Zhang prove logarithmic regret for network revenue management under degeneracy \cite{jiang2022degeneracy}.  This literature shares the LP-allocation flavor, but typically does not use uncensored demand to form counterfactual SAA objectives across prices.

\paragraph{Bandits, online optimization, and SAA.}
Badanidiyuru, Kleinberg, and Slivkins formulate bandits with knapsacks for learning under resource constraints \cite{badanidiyuru2013bandits}.  Chu et al. analyze linear contextual bandits, and Agarwal et al. give a reduction-based contextual-bandit framework \cite{chu2011contextual,agarwal2014taming}.  Agrawal and Kleinberg study continuum-armed bandits \cite{agrawal1995continuum,kleinberg2004nearly}.  Shalev-Shwartz and Hazan survey online learning and online convex optimization \cite{shalev2012online,hazan2019introduction}.  Agarwal et al. study stochastic convex optimization with bandit feedback \cite{agarwal2011stochastic}.  Birge and Louveaux and Shapiro, Dentcheva, and Ruszczynski provide standard stochastic-programming treatments \cite{birge1997introduction,shapiro2021lectures}; Kleywegt, Shapiro, and Homem-de-Mello analyze sample-average approximation \cite{kleywegt2002sample}; and Bonnans and Shapiro develop perturbation analysis for optimization problems \cite{bonnans2013perturbation}.  Our proof combines these themes, but the empirical scenarios are collected online and then algebraically translated across counterfactual prices.

\paragraph{Counterfactual feedback and deployment issues.}
Li et al. study unbiased offline evaluation from logged contextual-bandit data \cite{li2011unbiased}.  Dud\'ik, Langford, and Li propose doubly robust policy evaluation and learning \cite{dudik2011doubly}.  Bottou et al. frame counterfactual reasoning for computational advertising systems \cite{bottou2013counterfactual}.  Our counterfactual step is different: it is an algebraic demand-translation identity, not inverse-propensity evaluation.  Finally, fairness in price discrimination and dynamic pricing has been studied by Cohen, Elmachtoub, and Lei and by Cohen, Miao, and Wang \cite{cohen2022price,cohen2021dynamic}, and Spiliotopoulou and Conte study fairness ideals in inventory allocation \cite{spiliotopoulou2022fairness}. Meanwhile, Xu, Qiao, and Wang proposes a double fairness scheme under stochastic pricing strategies \cite{xu2023doubly}. We do not impose fairness constraints, but these works point to natural deployment-oriented extensions.

\subsection{Scope Clarifications and Potential Extensions}
\label{app:more_discussion}
We close with scope clarifications that also point to natural extensions of the model.

\paragraph{Uncensored feedback.}
The assumption of uncensored feedback matches preorder, quoting, marketplace logs, and demand-reporting environments, but not lost-sales retail where stockouts hide unserved demand.  The counterfactual translation identity relies on observing the demand vector itself, not only fulfilled sales.

\paragraph{Computation.}
The exact LCB minimization in the regret theorem isolates the statistical learning argument.  \Cref{prop:poly-additive-oracle} and \Cref{cor:approx-oracle} give a certified additive implementation for rational-polytope inventory sets by discretizing only price and solving fixed-price LPs.  This is polynomial-time for inverse-polynomial accuracy, but it is not intended as a large-scale practical solver.

\paragraph{Dimension dependence.}
The bounds match in their dependence on $T$ up to logarithmic factors, while the optimal dimension dependence remains open.  The diagonal lower bound shows that a linear dependence on $\min\{m,n\}$ is unavoidable; the current upper bound has a larger dimension factor due to direct grid-based SAA and coordinatewise plug-in error control.  This establishes the optimal horizon dependence, but does not rule out faster rates in restricted subclasses with slopes bounded away from zero.

\paragraph{Beyond linear demand.}
Linear demand makes counterfactual translation exact and cancels the intercept.  Nonlinear demand or censored observations would require new uniform error-control tools.

\paragraph{Societal concerns.}
If used with customer- or segment-level demand logs, pricing and allocation policies may raise privacy, fairness, and disparate-service concerns, especially under heterogeneous fulfillment costs.  These considerations are outside the current regret model and would require additional constraints or objectives.

\bigskip

\end{document}